\documentclass{article} 
\usepackage{iclr2027_conference,times}

\usepackage{amsmath,amsfonts,bm}

\def\eqref#1{equation~\ref{#1}}

\def\1{\bm{1}}

\DeclareMathAlphabet{\mathsfit}{\encodingdefault}{\sfdefault}{m}{sl}
\SetMathAlphabet{\mathsfit}{bold}{\encodingdefault}{\sfdefault}{bx}{n}

\usepackage{graphicx}
\usepackage{booktabs}
\usepackage{amsmath}
\usepackage{amssymb}
\usepackage{algorithm}
\usepackage{algpseudocode}
\usepackage[table]{xcolor}
\usepackage{hyperref}
\usepackage{url}
\usepackage{tabularx}
\usepackage[capitalise]{cleveref}
\usepackage{arydshln}
\usepackage{wrapfig,dsfont,pifont,caption,fix-cm,textcomp}
\usepackage[most]{tcolorbox} 

\definecolor{rowblue}{RGB}{220, 230, 242}
\definecolor{deltaup}{RGB}{0, 140, 0}
\definecolor{deltadn}{RGB}{200, 0, 0}
\definecolor{verylightgray}{rgb}{0.97, 0.97, 0.97}

\newcommand{\up}[1]{{\scriptsize\textcolor{deltaup}{+#1}}}
\newcommand{\dn}[1]{{\scriptsize\textcolor{deltadn}{-#1}}}
\newcommand{\xmark}{\textcolor{red}{\ding{55}}}

\newcommand{\ours}{VQS}

\title{Program-Verified Self-Evolution for Vision-Language Models}

\iclrfinalcopy
\author{Ahmed Heakl$^{1,2}$,
   Sungik Choi$^{1}$, 
   Moontae Lee$^{1,4}$, 
  Salman Khan$^{2,3}$ \\
  $^{1}$LG AI Research, $^{2}$MBZUAI, $^{3}$Australian National University,$^{4}$University of Illinois at Chicago \\
  \url{https://github.com/ahmedheakl/VQS}
}

\begin{document}

\maketitle

\begin{abstract}
Self-evolving vision-language models train on questions they generate from unlabeled images. Since these questions have no gold answers, prior methods label them by majority vote over sampled answers or by a model judge. In a human evaluation, we find that 24\% of majority-vote labels and 18\% of model-judge labels produced during self-evolution are wrong. To address this problem, we present Verifiable QA Generation for Self-Evolving Models (VQS), which changes how the model judges answers. Instead of voting on an answer, the model parses each image into a structured record, such as a scene graph, a chart table, or a diagram graph. Fixed programs then write a question from the record and compute its answer. The model still acts as a visual checker, but it only confirms the individual facts the program reads, one short claim at a time. These claim-level checks select the parser's training targets, so the parser also improves without labels. Human raters find 94\% of VQS answers correct, against 76\% for majority voting. Across ten benchmarks, VQS improves Qwen3-VL by up to 3.18 points at the 2B, 4B, and 8B scales and outperforms the strongest self-evolving baseline at each. Gains keep growing over three training rounds, reaching 3.84 points at 2B.
\end{abstract}

\section{Introduction}

\begin{wrapfigure}{r}{0.424\textwidth}
    \vspace{-1.8em}
    \centering
    \resizebox{\linewidth}{!}{
    \includegraphics[width=0.19\textwidth, , trim={22 20 22 20},clip]{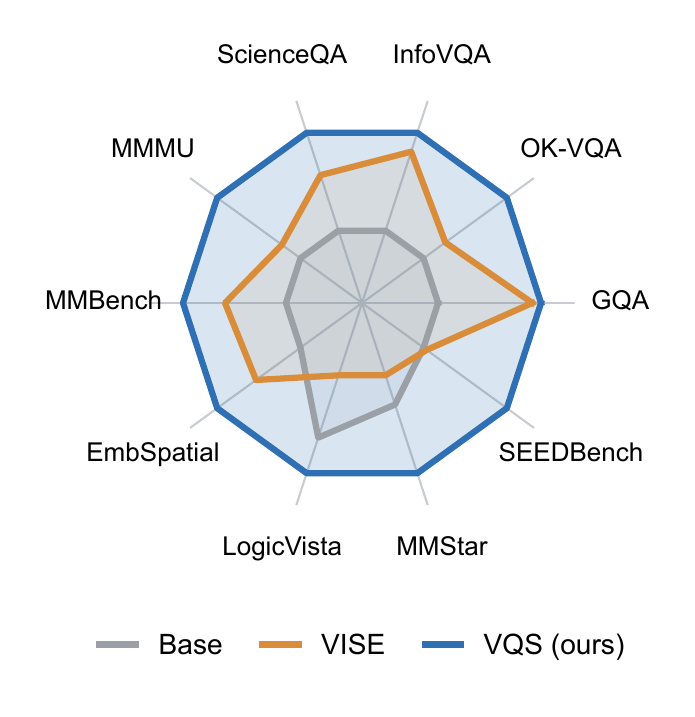}
    }
    \vspace{-1.6em}
    \caption{\textbf{VQS improves every benchmark.} Scores are for Qwen3-VL-2B. VISE loses on LogicVista and MMStar.}
    \vspace{-1.8em}
    \label{fig:teaser}
    
\end{wrapfigure}

A vision-language model can improve without human labels by writing the questions it trains on. Recent work does this by splitting one model into a proposer that asks and a solver that answers \citep{visplay, evolmm, ireasoner, rzero}. In each round, the proposer generates questions from unlabeled images, the solver answers them, and the solver is trained on the pairs that match a pseudo-label. Answer verification is the hardest step, since gold answers do not exist in this loop.

Standard reinforcement learning with verifiable rewards compares each answer with a gold label \citep{grpo, dapo}. However, self-generated questions have no gold label, so prior work falls back on proxy labels such as a majority vote over samples \citep{mmupt, ttrl, visplay}, agreement among reasoning steps \citep{ireasoner}, consistency under an image transformation \citep{vise}, or a model judge~\citep{mae, self-judge, self-reward}.

Prior works use proxies that accept many wrong answers. First, in our human evaluation of 500 generated questions (\cref{tab:human_evaluation}), 24\% of majority-vote labels and 18\% of model-judge labels are wrong. Second, the errors also grow as the loop runs. VisPlay's pseudo-label accuracy falls from 72\% to 61\% over three rounds \citep{visplay}, and R-Zero's falls from 79\% to 63\% \citep{rzero}. Third, prior work reaches the same conclusion independently, showing that majority-vote self-reward eventually favors confident but incorrect answers and collapses \citep{srt}.

\begin{figure}[t]
    \centering
    \includegraphics[width=\linewidth]{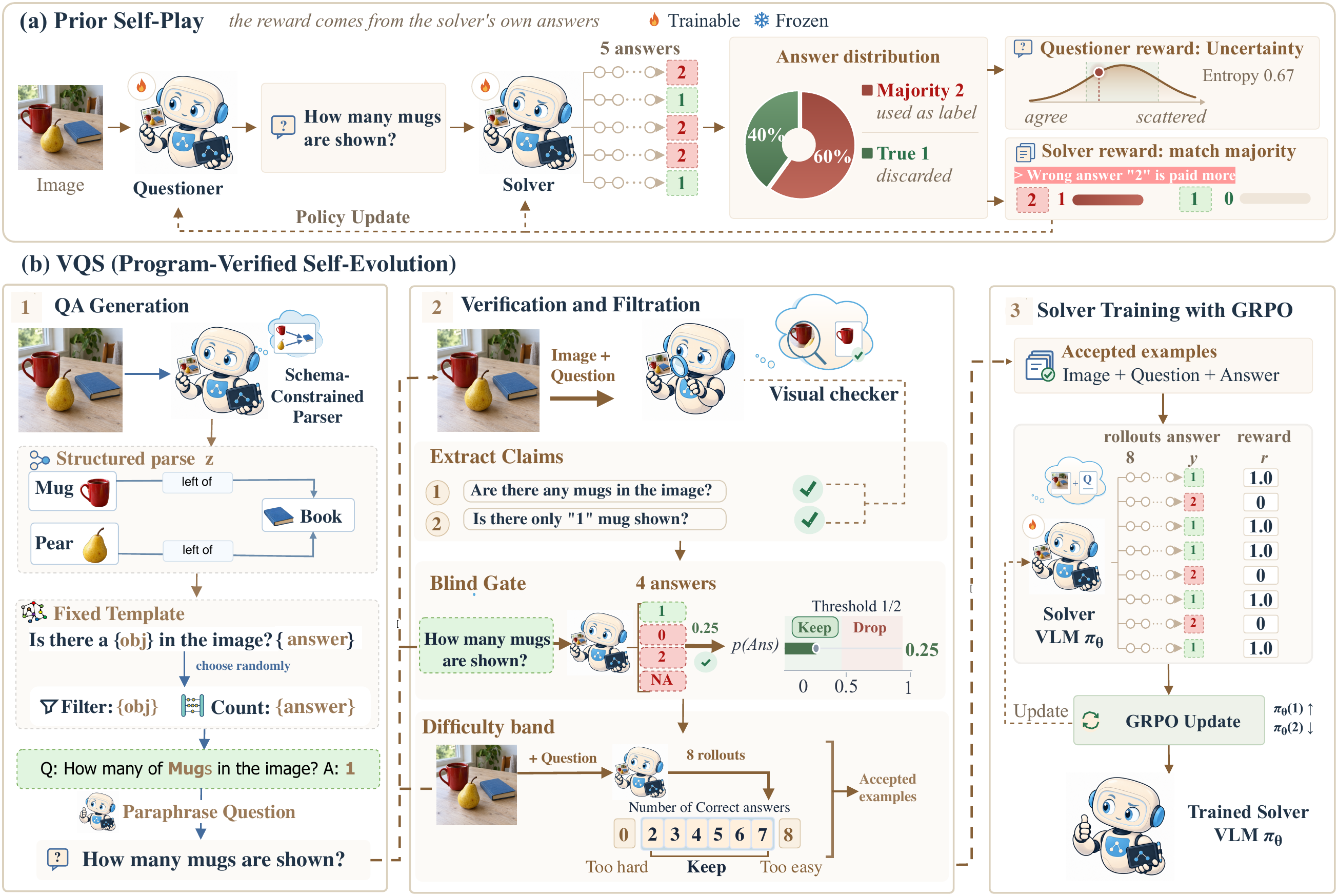}
    \vspace{-1.5em}
    \caption{\textbf{Majority-vote labels vs.\ computed answers.} \textbf{(a)} In prior self-play, the solver's majority answer becomes the label. Here three of five answers say 2 mugs, so the wrong answer is rewarded and the correct answer \textcolor{deltaup}{``1"} is discarded. \textbf{(b)} VQS computes the answer instead. \textbf{(1)} A schema-constrained parser turns the image into a structured parse. A fixed template reads the parse (filter mugs, then count) to write a question and compute its answer \textcolor{deltaup}{``1"} and the parser then paraphrases the question. \textbf{(2)} A visual checker verifies each claim behind the answer, a blind gate drops questions answerable without the image, and a difficulty band keeps those the solver gets right in some but not all of 8 rollouts. \textbf{(3)} GRPO rewards each rollout by exact match against the computed answer, so training reinforces the correct answer \textcolor{deltaup}{``1"} and suppresses the wrong one \textcolor{deltadn}{``2"}.}
    \label{fig:vqs_overview}
    \vspace{-1.1em}
\end{figure}

We present VQS (\textbf{V}erifiable \textbf{Q}A Generation for \textbf{S}elf-Evolving Models), a framework that computes answers from verified visual facts rather than estimating them from model outputs. As illustrated in \cref{fig:vqs_overview}, a parser first converts an input image into a navigable structure, such as a scene graph for a photograph, a structured table for a chart, or a set of nodes and directed edges for a diagram. Fixed templates (i.e., deterministic programs) then traverse these structures to generate questions alongside their computed answers. To catch wrong answers, a checker verifies the atomic visual facts behind an answer one at a time. We then train the solver on these verified QA pairs. These claim-level checks select the parser's training targets by choosing the most precise of $K$ parses, so the parser also improves without labels. A single model is used for the parser, solver, and checker.

Our contributions are: (1) VQS converts photographs, charts, diagrams, and infographics into structured records, then generates questions and computes answers through deterministic programs. (2) A label-free algorithm improves the parser by selecting its training set from candidate parses with claim-level visual verification. (3) VQS improves the average over ten benchmarks by up to \textbf{3.18} points (\cref{fig:teaser}) at the 2B, 4B, and 8B scales and beats the strongest self-evolving method at every scale. (4) VQS keeps improving over three training cycles on the same unlabeled images by generating new QA pairs from templates each cycle, raising the 2B gain from 3.18 to \textbf{3.84} points.

\section{Related work}

\textbf{Self-evolving vision-language models.} R-Zero \citep{rzero} uses a questioner and a solver trained against each other. VisPlay \citep{visplay} trains a questioner with a reward that peaks when the solver is maximally uncertain. EvoLMM \citep{evolmm} rewards the solver by how often its answer is the majority among five samples. iReasoner \citep{ireasoner} adds agreement between reasoning steps. Vision-Zero \citep{visionzero} turns the loop into a checkable game. VISE \citep{vise} drops the two-role setup and rewards the model for predicting the same box under a known image transformation. Without gold answers, all of these methods take their reward from the model's outputs, and this reward can become less accurate as training goes on. For example, VisPlay's estimated pseudo-label accuracy falls from 72\% to 61\% over three rounds as its questioner gets stronger, and R-Zero's falls from 79\% to 63\%.

\textbf{Programs that write questions.} CLEVR \citep{clevr} attaches a functional program to every question and runs it on the scene graph it rendered from. GQA \citep{gqa} scales this to real photographs using human-annotated graphs. ProVision \citep{provision} and SpatialVLM \citep{spatialvlm} run generators over a \emph{model-produced} parse at very large scale, but both produce fine-tuning data and neither verifies a generated claim.


\begin{figure}[t]
\centering
\includegraphics[width=\linewidth]{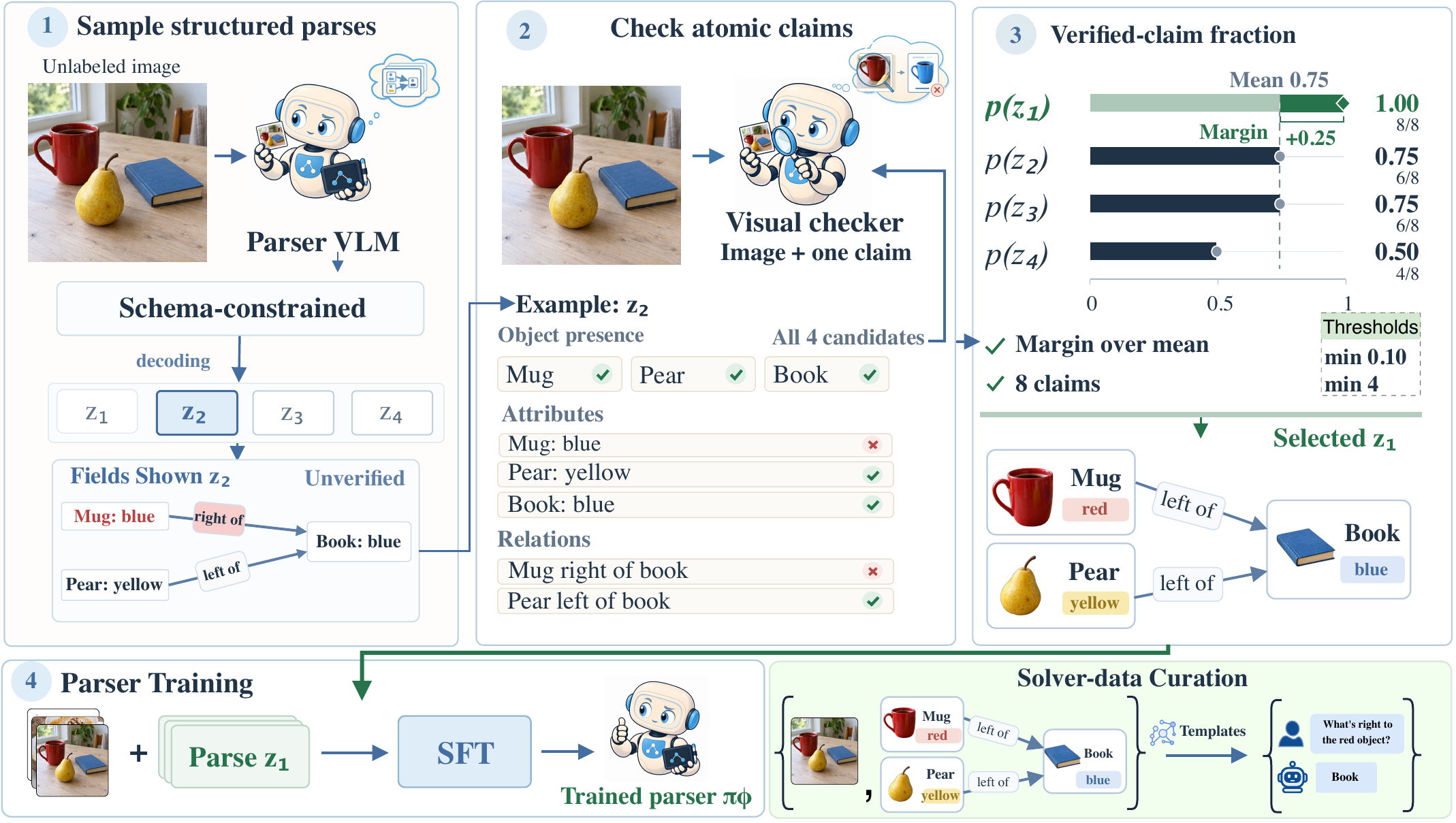}
\vspace{-1.5em}
\caption{\textbf{Label-free parser training.} \textbf{(1)} The parser samples $K$ structured parses of an unlabeled image under a schema-constrained decoder. \textbf{(2)} A visual checker verifies each parse one claim at a time. In the example, $z_2$ wrongly calls the mug blue and places it to the right of the book, so it passes 6 of 8 claims. \textbf{(3)} Each parse is scored by its fraction of verified claims. We keep the best parse $z_1$ (8/8) only because it beats the mean of all four parses (0.75) and contains at least $m{=}4$ claims. \textbf{(4)} The kept parses train the parser with SFT. The trained parser then feeds the templates that write questions and compute their answers for the solver.}
\label{fig:parser-pipeline}
\vspace{-1em}
\end{figure}


\section{Method}
\label{sec:method}

Below we describe how \ours{} builds the solver's training data, trains the solver, and trains the parser. In each training cycle, the parser is trained first. \Cref{alg:vqs} gives the full procedure.

\textbf{Setup.} The only training data is a set of unlabeled images $\mathcal{X}$. We initialize two models from the same pretrained model: (1) the \textbf{parser} $\pi_\phi$, which maps an image to a structured description that we call a \emph{parse}, and (2) the \textbf{solver} $\pi_\theta$, which answers questions about the image. The parser also serves as the checker and the text-only model described below (details in \cref{sec:setup}). A third component, the \textbf{program library} $\mathcal{T}$, is a fixed set of templates written once by hand and never trained (See \ref{app:examples}).

\begin{algorithm}[t]
\caption{\ours{} (one training cycle)}
\label{alg:vqs}
\begin{algorithmic}[1]
\Require unlabeled images $\mathcal{X}$, templates $\mathcal{T}$, pretrained weights $\pi_0$
\State $\phi, \theta \gets \pi_0$ \Comment{parser and solver start from the same model}
\Statex \textbf{Train the parser (label-free best-of-$K$)}
\For{$x \in \mathcal{X}$}
  \State sample $z^{(1..K)} \sim \pi_\phi(\cdot \mid x)$ under the schema
  \State score each $z^{(k)}$ by its fraction of verified claims $P(z^{(k)})$ \Comment{\cref{eq:bestofk}}
  \State keep $(x, z^\star)$ if it beats the mean of all $K$ parses by $\delta$ and has $\ge m$ claims \Comment{\cref{eq:keep_condition}}
\EndFor
\State fine-tune $\phi$ on the kept targets with cross-entropy loss
\Statex \textbf{Build the solver training data}
\For{$x \in \mathcal{X}$}
  \State $z \gets \pi_\phi(x)$, decoded greedily under the schema
  \State for each applicable template $t$: $(q, a, h) \gets t(z)$, then paraphrase $q \to \tilde q$
  \State \textbf{drop} the question unless it passes the fact-check, the blind gate, and the difficulty band \Comment{\cref{eq:band}}
\EndFor
\Statex \textbf{Train the solver}
\For{each GRPO step}
  \State draw $n$ rollouts per question and score each against $a$ \Comment{\cref{eq:grpo}}
  \State normalize rewards within each group and update $\theta$ with GRPO \Comment{\cref{eq:ppo}}
\EndFor
\State \Return $\pi_\theta$ \Comment{$\pi_\theta$ can re-parse the images and start the next cycle}
\end{algorithmic}
\vspace{-0.5em}
\end{algorithm}

\begin{figure}[t]
    \centering
    \includegraphics[width=\linewidth]{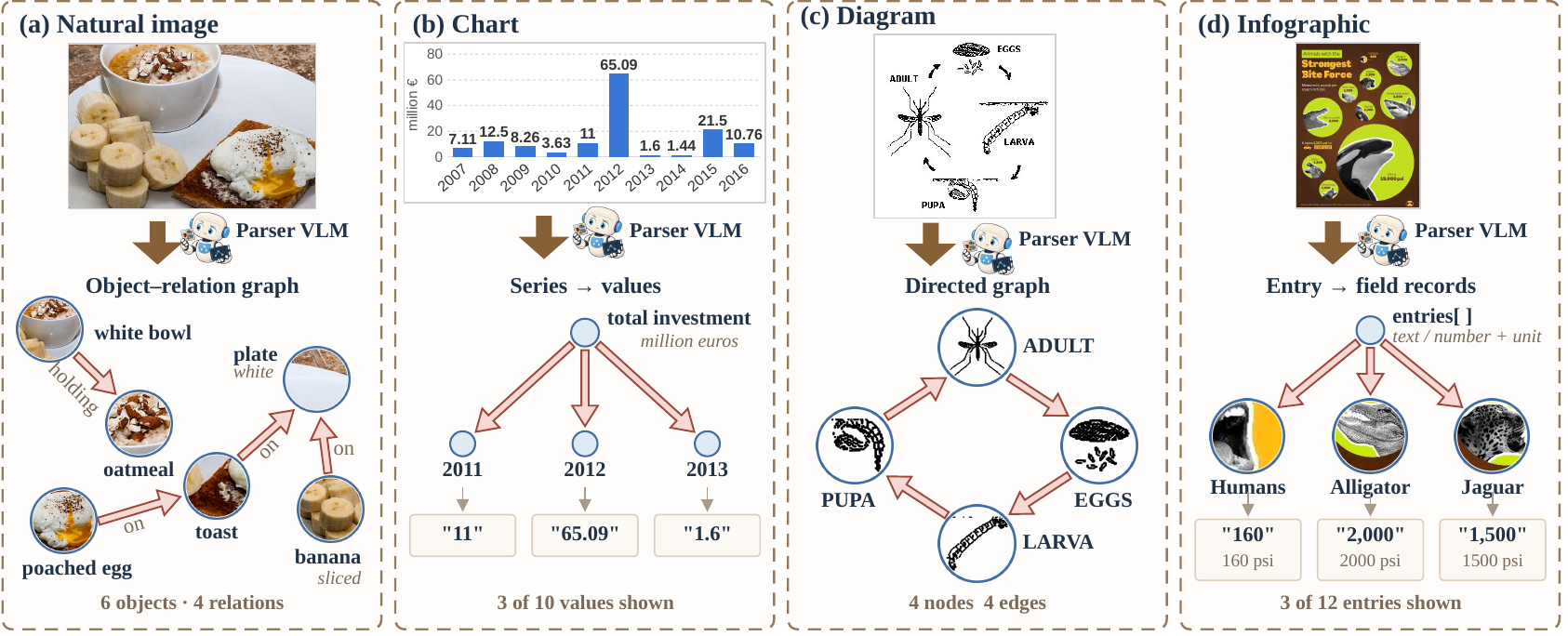}
    \caption{\textbf{Examples of structured scene parsing.} The parser converts natural images, charts, diagrams, and infographics into domain-specific JSON records that preserve spatial relations, chart values and ticks, graph structure, and verbatim strings for deterministic QA construction.}
    \label{fig:examples}
    \vspace{-1.5em}
\end{figure}

\textbf{From images to questions.} To build the solver's training data, the parser first maps each image $x$ to a parse $z$. Each parse must follow a schema $\mathcal{S}_{d(x)}$ fixed per domain $d(x) \in \{\text{photo}, \text{chart}, \text{diagram}, \text{infographic}\}$, and we decode it with a constrained decoder so that it is always well formed. A photo parse holds objects with names, attributes, boxes, and relations. A chart parse holds series of (category, printed value, numeric value) points and the axis ticks. A diagram parse holds nodes and directed arrows. An infographic parse holds entries with their text, number, and unit. \Cref{fig:examples} shows examples, and Appendix~\ref{app:prompts} gives the full schemas for each domain. 

Each applicable template $t \in \mathcal{T}$ then reads the parse $z$ and returns a question $q$, its answer $a$, and its hop count $h$, which is the number of operations the template performs: $(q, a, h) = t(z)$. Templates combine these operations: filter a set, read a field or attribute, count a set, take a maximum or minimum, follow a relation, rank, compare, do arithmetic, read a trend, count a node's arrows, and turn a box into a coarse position. Finally, the parser rewrites $q$ in natural wording, $\tilde q \sim \pi_\phi(\cdot \mid x, z, q)$, while the answer $a$ stays the same.

\textbf{Which questions enter the training data.} We keep a question only if it passes three checks. The \emph{fact-check} verifies the facts behind the answer. The template asserts one claim per field it reads, giving a set of atomic claims $\mathcal{C}(t,z)$, and a binary checker $v(c,x) \in \{0,1\}$ answers yes or no to each claim. The \emph{blind gate} removes questions that do not need the image. A text-only model $\pi_{\text{blind}}$ answers the question $J$ times without the image, $u_j \sim \pi_{\text{blind}}(\cdot \mid \tilde q)$, and we drop it if more than a fraction $\lambda$ of these answers are correct. The \emph{difficulty band} removes questions that are too easy or too hard. The solver about to be trained answers the question $n$ times, $y_i \sim \pi_\theta(\cdot \mid x, \tilde q)$, and we drop it if the solver is always right or always wrong, since all rollouts then get the same correctness score and give almost no learning signal (\cref{eq:grpo}). Formally, we keep the question only if
\begin{equation}
\underbrace{\textstyle\prod_{c \in \mathcal{C}(t,z)} \! v(c, x) = 1}_{\text{fact-check}}, \quad
\underbrace{\tfrac{1}{J}\!\sum_{j} \mathrm{eq}(u_j, a) \le \lambda}_{\text{blind gate}}, \quad
\underbrace{0 < \hat p < 1, \;\; \hat p = \tfrac{1}{n}\!\sum_{i} \mathrm{eq}(y_i, a)}_{\text{difficulty band}},
\label{eq:band}
\end{equation}
where $\mathrm{eq}(y, a) \in \{0,1\}$ is exact match between an answer $y$ and $a$.

\textbf{Solver Training.} The reward is an exact match with the computed answer, plus a small format term,
\begin{equation}
r(y; a) \;=\; 0.9 \cdot \mathrm{eq}(y, a) \;+\; 0.1 \cdot \mathrm{fmt}(y),
\label{eq:reward}
\end{equation}
where $\mathrm{fmt}(y) \in \{0,1\}$ checks that $y$ follows the required answer format. We train the solver with GRPO \citep{grpo} rather than SFT. GRPO learns on-policy, while SFT on template-written text may pull the model away from its distribution. For each question, GRPO normalizes the rewards of the $n$ rollouts within their group,
\begin{equation}
A_i \;=\; \frac{r_i - \mathrm{mean}(r_1 \dots r_n)}{\mathrm{std}(r_1 \dots r_n) + \varepsilon},
\qquad r_i = r(y_i; a),
\label{eq:grpo}
\end{equation}
and updates the solver with a clipped objective and a KL penalty against a frozen reference $\pi_{\text{ref}}$,
\begin{equation}
\mathcal{L}_{\text{solver}}(\theta) = -\,\mathbb{E}\Bigg[ \min\left(\frac{\pi_\theta(y_i)}{\pi_{\theta_{\text{old}}}(y_i)} A_i,\; \mathrm{clip}\left(\frac{\pi_\theta(y_i)}{\pi_{\theta_{\text{old}}}(y_i)}, 1-\epsilon_{\text{lo}}, 1+\epsilon_{\text{hi}}\right) A_i \right) \Bigg] \;+\; \beta \, \mathrm{KL}\!\left[\pi_\theta \,\|\, \pi_{\text{ref}}\right],
\label{eq:ppo}
\end{equation}
where $\epsilon_{\text{lo}}$ and $\epsilon_{\text{hi}}$ set the clipping range and $\beta$ weights the KL penalty.

\textbf{Parser Training.} The parser is trained first, before the solver data is built, and without labels (\cref{fig:parser-pipeline}). For each image $x$, we sample $K$ parses and score each by the fraction of its claims $\mathcal{F}(z)$, one per field, that the checker confirms:
\begin{equation}
z^{(1..K)} \sim \pi_\phi(\cdot \mid x), \qquad P(z) = \tfrac{1}{|\mathcal{F}(z)|} \!\sum_{c \in \mathcal{F}(z)}\! v(c, x), \qquad
z^\star = \arg\max_k P(z^{(k)}).
\label{eq:bestofk}
\end{equation}
We keep the best parse $z^\star$ only when it clearly beats the average of all $K$ parses, so that each kept target carries a real learning signal. If all $K$ parses score the same, the chosen parse is no better than what the parser already produces. We also require the best parse to have at least $m$ claims, since small parses usually get inflated verification scores:
\begin{equation}
\text{keep } (x, z^\star) \text{ iff } P(z^\star) - \tfrac{1}{K}\!\sum_k P(z^{(k)}) \ge \delta
\;\text{ and }\; |\mathcal{F}(z^\star)| \ge m.
\label{eq:keep_condition}
\end{equation}
\vspace{-8pt}
 
Finally, we fine-tune the parser on the kept targets $(x, z^\star)$ with the standard cross-entropy loss.

\section{Experimental setup}
\label{sec:setup}

\textbf{Models and data.} We train Qwen3-VL-2B, 4B and 8B Instruct~\citep{qwen3-vl}. We take the training images from \citet{vise} and discard their questions, answers, and labels, so no human annotation is used at any stage.

\textbf{Benchmarks.} We use lmms-eval~\citep{lmmseval} to evaluate on 10 benchmarks: GQA~\citep{gqa}, OK-VQA~\citep{okvqa}, InfoVQA~\citep{infovqa}, ScienceQA~\citep{scienceqa}, MMMU~\citep{mmmu}, MMBench~\citep{mmbench}, EmbSpatial~\citep{embspatial}, LogicVista~\citep{logicvista}, MMStar~\citep{mmstar}, SEEDBench~\citep{seedbench}. We use the default hyperparameters for each benchmark. 

\textbf{Implementation Details.} We used 8xAMD MI210 cards for training and evaluation. Every reinforcement learning run uses GRPO via EasyR1~\citep{easyr1} and the parser is fine-tuned with LLaMA-Factory~\citep{llamafactory}. For parsing, we sample each image $K{=}4$ parses at temperature 1.0 and top-$p$ 0.95, up to 2048 tokens, under a per-domain JSON schema the decoder enforces, and score each by the fraction of its claims the checker confirms. We keep the best parse only if it beats the mean of all $K$ parses by $\delta{=}0.10$ and asserts at least $m{=}4$ claims, which turned 12{,}000 images into 4{,}792 targets. The parser is then fully fine-tuned on those targets with the vision tower frozen, per-device batch 4 with 8 gradient accumulation steps, learning rate $1\mathrm{e}{-5}$, 2 epochs, 284 steps. At data-curation time it decodes greedily. The fact-checker $v$ is the same as the parser answering yes or no to one claim at a time, greedy, with four new tokens. The blind gate $\pi_{\text{blind}}$ is the same as the parser with no image. It generates $J{=}4$ answers and drops a question if more than a fraction $\lambda{=}0.5$ of them are correct. The difficulty band runs the base solver at the training settings, 8 rollouts at temperature 1.0 with a 256-token budget, and keeps a question only when it is answered correctly at least once and not every time. We train the solver with a curriculum that sorts each template family's questions by hop count $h$, from fewest to most (\cref{tab:revision_question_design}).

\section{Results}
\label{sec:main}

We answer five questions: (1) Does VQS beat prior self-evolving methods, and how do its gains change with model size? (2) How many generated questions pass the filters, and are their answers correct? (3) Does label-free training make the parser more accurate? (4) Which filters and questions help the solver most? (5) Does the model keep improving over more training cycles?

\begin{table}[t]
\centering
\caption{\textbf{Main results across three scales.} Every model trains on unlabeled images only, with no human question, answer or gold label at any stage. Subscripts give the change against that backbone's own base, in \textcolor{deltaup}{green} when positive and \textcolor{deltadn}{red} when negative. \textit{VQS gives the largest average gain at all three scales}, +3.18 at 2B, +2.32 at 4B and +2.43 at 8B, against +1.01, +1.04 and +0.66 for the strongest published method. InfoVQA stands for infographicsVQA, SQA for ScienceQA, MMB for MMBench, ESB for EmbSpatial, and LogicV for LogicVista.}
\vspace{-0.5em}
\setlength{\tabcolsep}{3pt}
\resizebox{\textwidth}{!}{%
\begin{tabular}{lccccccccccc}
\toprule
\textbf{Method} & \textbf{GQA} & \textbf{OK-VQA} & \textbf{InfoVQA} & \textbf{SQA} & \textbf{MMMU} & \textbf{MMB} & \textbf{ESB} & \textbf{LogicV} & \textbf{MMStar} & \textbf{SEED} & \textbf{Avg.} \\
\midrule
\multicolumn{12}{l}{\textit{Qwen3-VL-2B-Instruct}} \\
\hdashline
Base & 58.25 & 40.76 & 69.02 & 79.42 & 38.92 & 74.48 & 68.54 & 35.04 & 55.62 & 72.16 & 59.22 \\
VisPlay & 58.65\up{0.40} & 41.15\up{0.39} & 69.96\up{0.94} & 80.74\up{1.32} & 39.27\up{0.35} & 74.52\up{0.04} & 68.56\up{0.02} & 34.93\dn{0.11} & 55.37\dn{0.25} & 71.62\dn{0.54} & 59.48\up{0.26} \\
Vision-Zero & 58.98\up{0.73} & 41.39\up{0.63} & 70.93\up{1.91} & 81.96\up{2.54} & 39.58\up{0.66} & 75.07\up{0.59} & 69.72\up{1.18} & 35.28\up{0.24} & 55.58\dn{0.04} & 71.53\dn{0.63} & 60.00\up{0.78} \\
EvoLMM & 59.01\up{0.76} & 38.03\dn{2.73} & 70.69\up{1.67} & 83.01\up{3.59} & 39.08\up{0.16} & 74.62\up{0.14} & 69.32\up{0.78} & 34.99\dn{0.05} & 55.50\dn{0.12} & 71.17\dn{0.99} & 59.54\up{0.32} \\
iReasoner & 59.13\up{0.88} & 38.13\dn{2.63} & 70.82\up{1.80} & 83.12\up{3.70} & 39.11\up{0.19} & 74.75\up{0.27} & 69.67\up{1.13} & 35.09\up{0.05} & 55.59\dn{0.03} & 71.25\dn{0.91} & 59.67\up{0.45} \\
VISE & 59.41\up{1.16} & 41.24\up{0.48} & 71.43\up{2.41} & 83.61\up{4.19} & 40.67\up{1.75} & 76.72\up{2.24} & 70.14\up{1.60} & 31.92\dn{3.12} & 55.02\dn{0.60} & 72.18\up{0.02} & 60.23\up{1.01} \\
\rowcolor{rowblue}
\textbf{VQS (ours)} & \textbf{59.52}\up{1.27} & \textbf{42.60}\up{1.84} & \textbf{72.01}\up{2.99} & \textbf{86.81}\up{7.39} & \textbf{46.78}\up{7.86} & \textbf{78.26}\up{3.78} & \textbf{71.54}\up{3.00} & \textbf{36.83}\up{1.79} & \textbf{57.04}\up{1.42} & \textbf{72.61}\up{0.45} & \textbf{62.40}\up{3.18} \\
\midrule
\multicolumn{12}{l}{\textit{Qwen3-VL-4B-Instruct}} \\
\hdashline
Base & 60.32 & 47.86 & 77.73 & 87.51 & 45.17 & 83.42 & 76.79 & 49.78 & 62.67 & 75.71 & 66.70 \\
VisPlay & 60.06\dn{0.26} & 48.04\up{0.18} & 77.96\up{0.23} & 87.30\dn{0.21} & 45.42\up{0.25} & 83.47\up{0.05} & 76.11\dn{0.68} & 49.34\dn{0.44} & 62.13\dn{0.54} & 74.83\dn{0.88} & 66.47\dn{0.23} \\
Vision-Zero & 61.30\up{0.98} & 48.41\up{0.55} & 78.28\up{0.55} & 88.46\up{0.95} & 46.27\up{1.10} & \textbf{84.83}\up{1.41} & 77.02\up{0.23} & 49.11\dn{0.67} & 62.45\dn{0.22} & 75.01\dn{0.70} & 67.19\up{0.49} \\
EvoLMM & 61.20\up{0.88} & 46.72\dn{1.14} & 78.58\up{0.85} & 88.46\up{0.95} & 46.02\up{0.85} & 84.21\up{0.79} & 76.94\up{0.15} & 49.57\dn{0.21} & 62.37\dn{0.30} & 74.77\dn{0.94} & 66.88\up{0.19} \\
iReasoner & 61.32\up{1.00} & 46.89\dn{0.97} & 78.76\up{1.03} & 88.62\up{1.11} & 46.13\up{0.96} & 84.34\up{0.92} & 76.98\up{0.19} & 49.65\dn{0.13} & 62.46\dn{0.21} & 74.84\dn{0.87} & 67.00\up{0.30} \\
VISE & 61.82\up{1.50} & \textbf{48.81}\up{0.95} & \textbf{81.45}\up{3.72} & 90.04\up{2.53} & 48.89\up{3.72} & 84.51\up{1.09} & 77.46\up{0.67} & 47.66\dn{2.12} & 61.73\dn{0.94} & 75.00\dn{0.71} & 67.74\up{1.04} \\
\rowcolor{rowblue}
\textbf{VQS (ours)} & \textbf{62.86}\up{2.54} & 48.19\up{0.33} & 80.23\up{2.50} & \textbf{93.11}\up{5.60} & \textbf{54.78}\up{9.61} & 83.93\up{0.51} & \textbf{77.80}\up{1.01} & \textbf{50.10}\up{0.32} & \textbf{63.29}\up{0.62} & \textbf{75.82}\up{0.11} & \textbf{69.01}\up{2.32} \\
\midrule
\multicolumn{12}{l}{\textit{Qwen3-VL-8B-Instruct}} \\
\hdashline
Base & 61.54 & 49.84 & 81.23 & 90.88 & 50.12 & 84.71 & 77.66 & 52.23 & 64.96 & 74.96 & 68.81 \\
VisPlay & 61.66\up{0.12} & 49.93\up{0.09} & 81.36\up{0.13} & 91.02\up{0.14} & 52.61\up{2.49} & 84.62\dn{0.09} & 78.24\up{0.58} & 51.88\dn{0.35} & 64.52\dn{0.44} & 74.34\dn{0.62} & 69.02\up{0.21} \\
Vision-Zero & 62.12\up{0.58} & 50.31\up{0.47} & 82.44\up{1.21} & 92.35\up{1.47} & 52.89\up{2.77} & 85.11\up{0.40} & 77.96\up{0.30} & 52.33\up{0.10} & 64.75\dn{0.21} & 74.33\dn{0.63} & 69.46\up{0.65} \\
EvoLMM & 61.94\up{0.40} & 50.18\up{0.34} & 82.02\up{0.79} & 91.96\up{1.08} & 51.56\up{1.44} & 85.02\up{0.31} & 77.51\dn{0.15} & 52.01\dn{0.22} & 64.59\dn{0.37} & 74.12\dn{0.84} & 69.09\up{0.28} \\
iReasoner & 62.02\up{0.48} & 50.24\up{0.40} & 82.15\up{0.92} & 92.11\up{1.23} & 51.83\up{1.71} & 85.09\up{0.38} & 77.82\up{0.16} & 52.09\dn{0.14} & 64.67\dn{0.29} & 74.18\dn{0.78} & 69.22\up{0.41} \\
VISE & \textbf{62.43}\up{0.89} & 50.61\up{0.77} & 82.83\up{1.60} & 92.81\up{1.93} & 52.69\up{2.57} & 85.46\up{0.75} & \textbf{78.62}\up{0.96} & 50.78\dn{1.45} & 64.17\dn{0.79} & 74.30\dn{0.66} & 69.47\up{0.66} \\
\rowcolor{rowblue}
\textbf{VQS (ours)} & 
61.97\up{0.43} & 
\textbf{53.00}\up{3.16} & 
\textbf{83.13}\up{1.90} & 
\textbf{94.65}\up{3.77} & 
\textbf{57.56}\up{7.44} & 
\textbf{85.82}\up{1.11} & 
78.49\up{0.83} & 
\textbf{54.24}\up{2.01} & 
\textbf{68.14}\up{3.18} & 
\textbf{75.39}\up{0.43} & 
\textbf{71.24}\up{2.43} \\
\bottomrule
\end{tabular}
}
\label{tab:main}
\vspace{-1.0em}
\end{table}

\textbf{Main Results.}
\Cref{tab:main} compares VQS with five self-evolving methods on ten benchmarks at three Qwen3-VL scales. Existing consistency-based baselines using majority voting or sample agreement show relatively little improvement, as they depend significantly on the base model performance. For example, EvoLMM and iReasoner raise ScienceQA, where the 2B base is already strong, but lower OK-VQA, where the base is weak. VisPlay stays close to the base scores on most benchmarks, because its questioner is rewarded for questions the solver is least sure about, and a majority vote is least reliable on such questions.

VQS gives the largest average gain at every scale (+3.18 over base at 2B, against +1.01 for the strongest baseline) and lifts all ten benchmarks at each. Its labels come from programs run on checked parses, not from the solver, so training can move the model toward answers it did not favour and rarely rewards a wrong one (\cref{tab:human_evaluation}). This helps most where the base is weak. At 2B, the majority-vote methods gain almost nothing on MMMU, while VQS gains more than twice as much as any baseline (+7.86 vs +1.75). MMMU and ScienceQA have the largest gains as they ask the model to read a chart, table, or figure and combine several facts, exactly what the multi-step templates train (\cref{tab:revision_family_gains}). VISE, the strongest baseline, lowers the scores on LogicVista and MMStar at every scale, while VQS raises both. Like MMMU and ScienceQA, both benchmarks contain many questions that require combining several facts. See \ref{app:model-families} for results on non-Qwen model families.

\begin{figure}[t]
\centering
\includegraphics[width=\linewidth]{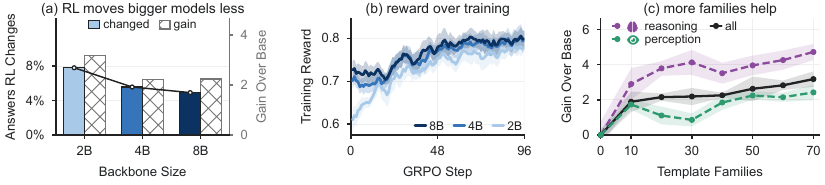}
\vspace{-1.5em}
\caption{\textbf{What scale and coverage change.} \textbf{(a)} Share of benchmark answers RL changes, measured per sample against each model's base. \textbf{(b)} Smoothed training reward. \textbf{(c)} 2B gain over base against the number of template families.}
\label{fig:scale}
\vspace{-1.5em}
\end{figure}

\textbf{Effect of model scale.}
Gains from VQS are largest at the smallest backbone and shrink with size, from $+3.18$ at 2B to $+2.43$ at 8B. All three models reach a training reward of about $0.80$ (\cref{fig:scale}b), so the larger models learn the generated task as well as the 2B model, but less of what they learn transfers to the benchmarks. Since an average gain cannot distinguish a model that barely changed from one that changed many answers in opposite directions, we also compare each run with its base sample by sample and count the benchmark answers that change. This share falls from $7.82\%$ at 2B to $5.58\%$ at 4B and $4.93\%$ at 8B (\cref{fig:scale}a), which matches the smaller gains of the larger models.

\begin{wraptable}{r}{0.4\linewidth}
    \vspace{-\intextsep}
    \centering
    \fontsize{8.5}{10.2}\selectfont
    \setlength{\tabcolsep}{3.5pt}
    \renewcommand{\arraystretch}{1.12}
    \rowcolors{2}{verylightgray}{white}
    \resizebox{\linewidth}{!}{
    \begin{tabular}{@{}lrr@{}}
        \toprule
        Answer source & Correct (\%) & Kept (\%) \\
        \midrule
        Templates + checker & \textbf{94.4\%} & 76.0 \\
        Majority vote & 76.4\% & 92.0 \\
        Model judge & 82.2\% & 90.8 \\
    \bottomrule
    \end{tabular}
    }
    \caption{\textbf{Human evaluation.} Correct is accuracy among retained questions.}
    \label{tab:human_evaluation}
    \vspace{-0.7\baselineskip}
\end{wraptable}

\textbf{Do the generated answers match the image?} Since each answer is computed from the parse, a parsing error can still produce a wrong answer, so human raters check the generated answers against the images. We compare three answer sources on 500 shared questions, with 125 per visual domain and coverage across template families. VQS executes templates on checked parses. Majority voting selects the most frequent of five solver answers and keeps it only if $p(y_{majority})>0.3$. A model judge selects an accepted answer from the same five candidates. Table~\ref{tab:human_evaluation} reports both correctness and retention: VQS reaches 94.4\% correctness while keeping 76\% of questions, compared with 76.4\% for majority voting and 82.2\% for the model judge.

\textbf{Does paraphrasing change the question?} The parser rewrites each template question in natural wording, but neither the checker nor the blind gate tests whether the rewrite still asks what the program computed. We check this on 437 questions from 70 template families. The final question still matches its program in 99.8\% of cases at 2B, 100\% at 4B, and 99.1\% at 8B. All four 8B errors swap the two years in a yes/no comparison, so the stored answer becomes wrong.

\begin{wraptable}{r}{0.54\linewidth}
    \vspace{-1.0em}
    \centering
    \fontsize{8.5}{10}\selectfont
    \setlength{\tabcolsep}{3.5pt}
    \rowcolors{2}{verylightgray}{white}
    \resizebox{\linewidth}{!}{
    \begin{tabular}{@{}lrrr@{}}
        \toprule
        \textbf{Parser target} & \textbf{Parse Prec.} & \textbf{Answer Acc.} & \textbf{Solver Acc.} \\
        \midrule
        Untrained & 80.8 & 86.8 & 58.0 \\
        Random & 82.1 & 91.0 & 61.5  \\
        Whole parse & 85.0 & 94.0 & 62.0 \\
        Per claim & \textbf{87.6} & \textbf{96.1} & \textbf{62.4} \\
        \bottomrule
    \end{tabular}
    }
    \vspace{-0.5em}
    \caption{\textbf{Parser target selection.} Parse precision, correct answer accuracy, and solver accuracy (\%).}
    \label{tab:revision_selection}
    \vspace{-8pt}
\end{wraptable}%

\textbf{Parser training selection.}\label{sec:revision_parser_selection}
We test how to pick the parser's training target from its $K$ samples without labels. We compare three rules against the untrained parser: a random parse, the parse a checker accepts as a whole with one yes or no, and the parse with the highest fraction of verified claims. We rate parse precision and answer accuracy by hand on 150 samples. We also train a 2B solver on the questions built from each parser and report its average over the ten benchmarks. \Cref{tab:revision_selection} shows that per-claim checking is best on all three measures. Over whole-parse checking, it raises parse precision from 85.0 to 87.6 and answer accuracy from 94.0 to 96.1. Solver accuracy follows the same order. Training the parser with any rule lifts it from 58.0 to at least 61.5, and the three rules differ by less than one point. Questions from the untrained parser even drop the solver below the 2B base (59.22, \cref{tab:main}), likely because 13.2\% of their answers are wrong. One reason per-claim checking works better is that a single yes or no cannot tell a parse with one error from a parse with five. Hence, when many samples tie, the retained parse is often chosen by chance. In contrast, the fraction of verified claims gives a graded score that separates them. Another reason is that a checker judges one short claim more reliably than a long structure~\citep{factscore, faithscore}, in line with findings that checking each fact separately reduces errors~\citep{cove}.

\begin{wrapfigure}{r}{0.4\linewidth}
\vspace{-1.2em}
\centering
\includegraphics[width=\linewidth]{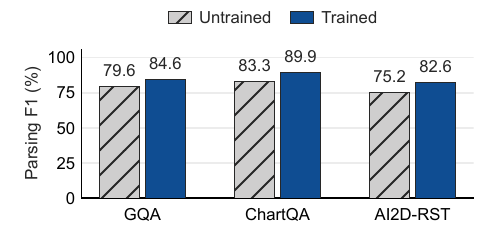}
\vspace{-2.3em}
\caption{\textbf{Parser Structured-F1 Score.}}
\label{fig:revision_parser_accuracy}
\vspace{-1.2em}
\end{wrapfigure}

\textbf{Does label-free training improve the parser?}
Every VQS answer is computed from the parse, so a parsing error becomes a wrong answer. The parser is also trained on the checker's verdicts, so it could learn to please the checker without becoming more accurate. We therefore evaluate the parser against human annotations. We measure F1 against GQA scene graphs~\citep{gqa}, ChartQA tables~\citep{chartqa}, and AI2D-RST diagram annotations~\citep{ai2drst}. \Cref{fig:revision_parser_accuracy} compares F1 by domain. Label-free training improves F1 in all three domains, by 5.0 points on GQA, 6.6 on ChartQA, and 7.4 on AI2D-RST. The largest gain comes on diagrams, where the untrained parser is weakest (75.2), although diagrams remain the hardest domain after training (82.6).

\begin{figure}[t]
\centering
\includegraphics[width=\linewidth]{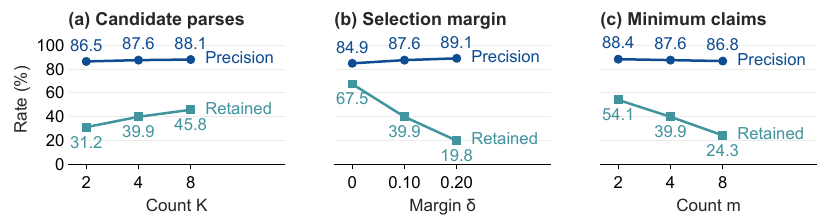}
\vspace{-1.5em}
\caption{\textbf{Parser target selection: quality and coverage.} Each panel varies one setting around $(K,\delta,m)=(4,0.10,4)$. Parse precision is the percentage of correct visual claims after training. Images retained is the percentage of source images yielding a selected training target.}
\label{fig:revision_parser_sweeps}
\vspace{-1.5em}
\end{figure}

\textbf{Parser hyperparameter selection.} We also vary the candidate count $K$, the selection margin $\delta$, and the minimum claim count $m$ (\cref{fig:revision_parser_sweeps}). Raising $K$ or $\delta$ improves precision, but only slightly, so we choose a middle setting that keeps computation affordable. Raising $m$, however, slightly lowers precision. Parses with more claims can produce more questions, but longer parses are more likely to contain hallucinated content \citep{vista}.

\begin{wrapfigure}{r}{0.4\linewidth}
\vspace{-2.1em}
\centering
\includegraphics[width=\linewidth]{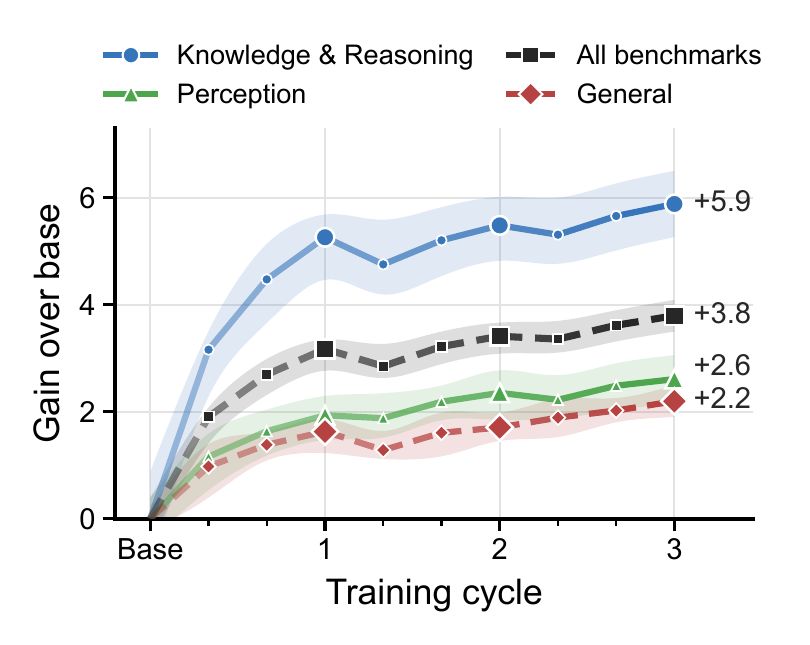}
\vspace{-2.2em}
\caption{\textbf{Gains across cycles.} Average gain over base per benchmark category for Qwen3-VL-2B.}
\label{fig:cycles}
\vspace{-1.5em}
\end{wrapfigure}

\textbf{Does VQS keep improving over several rounds?}
All results so far (\cref{tab:main}) come from a single round, or training cycle, in which the parser and the solver are each trained once. Here we test whether more cycles help the solver. At the start of each new cycle, we merge the LoRA weights from the previous solver cycle, generate new QA pairs from our templates on the same images, and combine them with the QA pairs from earlier cycles. We then sample 8 answers to each question, and compute $s$ as the fraction of these answers that are correct. We weight each question by $w(s)=\exp(-((s-0.5)/0.15)^2)$, give zero weight to questions that are always or never solved, and resample 8K training items with these weights. As a result, questions solved about half the time appear most often. Finally, we train the model from the previous cycle on these items for 96 GRPO steps, with the same settings as all other runs. The average gain over base grows with every cycle (\cref{fig:cycles}), from 3.18 points after cycle 1 to 3.41 and 3.84 after cycles 2 and 3. The third cycle adds more than the second (0.43 vs 0.23 points), so the curve has not flattened. As with a single cycle, knowledge and reasoning benchmarks gain the most, 5.94 points after three cycles, led by ScienceQA ($+8.53$) and MMMU ($+8.44$). Perception and general benchmarks gain 2.62 and 2.24 points. Nine of the ten benchmarks improve in every cycle, and the only exception, MMStar, peaks after cycle 1 but still ends 0.93 points above base.

\begin{wraptable}{r}{0.4\linewidth}
    \centering
    \fontsize{8.5}{10.2}\selectfont
    \setlength{\tabcolsep}{3.0pt}
    \rowcolors{2}{verylightgray}{white}
    \resizebox{\linewidth}{!}{
    \begin{tabular}{lrr}
    \toprule
    Pipeline & Solver Acc. & Parser Prec. \\
    \midrule
    Untrained parser & 58.0 & 80.8 \\
    
    Full VQS & \textbf{62.4} & \textbf{87.6} \\
    \xmark~ constrained decoding & 61.9 & 85.3 \\
    \xmark~ fact-checker & 60.9 & 87.6 \\
    \xmark~ blind gate & 61.8 & 87.6 \\
    \xmark~ difficulty band & 62.1 & 87.6 \\
    \bottomrule
    \end{tabular}
    }
    \caption{\textbf{Component ablations at 2B.} }
    \label{tab:components}
    \vspace{-0.7\baselineskip}
\end{wraptable}

\textbf{Which filters improve solver accuracy?}\label{sec:revision_filters}
\Cref{tab:components} removes one component at a time at 2B. The filters act after parsing, so removing one changes solver accuracy but not parser precision. Removing parser training causes the largest drop, lowering parser precision from 87.6 to 80.8 and solver accuracy from 62.4 to 58.0. Removing the fact-checker causes the second-largest drop, lowering solver accuracy to 60.9 while parser precision stays at 87.6, so unchecked answers alone cost 1.5 points. Each remaining component costs 0.3 to 0.6 points.

\begin{wraptable}{r}{0.35\linewidth}
    \vspace{-1.3em}
    \centering
    \footnotesize
    \setlength{\tabcolsep}{3pt}
    \rowcolors{2}{verylightgray}{white}
    \resizebox{\linewidth}{!}{
    \begin{tabular}{llr}
        \toprule
        \textbf{Rank} & \textbf{Family} & \textbf{Gain} \\
        \midrule
        \textcolor{green!50!black}{{\small$\blacktriangle$}}\ 1 & Chart differences & 7.4 \\
        \textcolor{green!50!black}{{\small$\blacktriangle$}}\ 2 & Diagram connectivity & 5.8 \\
        \textcolor{red!80!black}{{\small$\blacktriangledown$}}\ 1 & Object existence & -0.2 \\
        \textcolor{red!80!black}{{\small$\blacktriangledown$}}\ 2 & Attribute lookup & 0.4 \\
        \bottomrule
    \end{tabular}
    }
    \vspace{-1em}
    \caption{\textbf{Per-family gains.} The two largest and two smallest.}
    \label{tab:revision_family_gains}
    \vspace{-1.5em}
\end{wraptable}%

\textbf{Which families gain the most?} We measure the solver's gain from each template family, and \cref{tab:revision_family_gains} lists the two families with the largest gains and the two with the smallest. The largest gains come from families that combine several reads, such as subtracting two chart values or following diagram arrows. Families that need a single read, such as object existence and attribute lookup, barely change, likely because the base model already answers them reliably. This matches the main results, where reasoning-heavy benchmarks such as MMMU and ScienceQA gain the most. Lastly, \cref{fig:scale}c shows that adding template families keeps raising the gain over base. The average gain grows from 1.9 points with 10 families to 3.18 with all 70, and it is still rising. Reasoning benchmarks gain more than perception benchmarks at every family count, reaching 4.7 against 2.4 points.

\begin{wraptable}{r}{0.48\linewidth}
\centering
\fontsize{8.5}{10.5}\selectfont
\setlength{\tabcolsep}{3pt}
\renewcommand{\arraystretch}{1.08}
\begin{tabular}{@{}llr@{}}
\toprule
Comparison & Setting & Solver acc. \\
\midrule
Family difficulty & Easy & 60.8 \\
 & Middle & \textbf{62.4} \\
 & Hard & 61.1 \\
\midrule
Generator & Fixed templates  & \textbf{62.4} \\
 & Generated programs & 61.4 \\
\midrule
Curriculum & Shuffled  & 61.3 \\
 & Global hops & 61.7 \\
 & Within-family hops & \textbf{62.4} \\
\midrule
Difficulty band & Initial selection  & \textbf{62.4} \\
 & Refreshed selection & 62.3 \\
\bottomrule
\end{tabular}
\caption{\textbf{Question design at 2B.} Each block changes one choice at an equal training budget. Solver accuracy averages the 10 benchmarks in \cref{tab:main}. Bold marks the VQS default.}
\label{tab:revision_question_design}
\vspace{-10pt}
\end{wraptable}

\textbf{Which questions are most useful for training?}\label{sec:revision_question_design}
Beyond a correct answer, a question helps the solver only if it needs the image and the current solver answers it correctly sometimes, but not always. We test four design choices at equal training budgets (\cref{tab:revision_question_design}). For difficulty, we split template families into easy, middle, and hard thirds by the base solver's accuracy and train on each third alone. For the generator, we replace fixed templates with model-written questions and programs on the same parses, following PROVE \citep{prove}. Both generators use the same checkers, and human raters check whether each question asks what its program computes. For the curriculum, we compare shuffled order with ordering by hop count, across all questions or within each family. For the difficulty band, we compare selecting questions once with the base solver against reselecting them with the current solver during training. Middle families work best (62.4), ahead of hard (61.1) and easy (60.8), which matches prior findings that questions of intermediate difficulty give the most learning signal \citep{odf, pcl}. Generated programs lose 1.0 point because only 85.3\% of their questions match their program. Ordering by hop count helps, and ordering within each family works best, raising accuracy from 61.3 (shuffled) to 62.4. Reselecting during training brings no gain (62.3), so one selection with the base solver is enough.

\vspace{-0.5em}
\section{Conclusion}
\vspace{-0.5em}

In this work, we show that self-evolving vision-language models can generate hallucinated questions and answers that still pass consistency-based verifiers. In our human evaluation, 24\% of majority-vote labels and 18\% of model-judge labels are wrong. To enable principled generation and verification, we propose VQS, a template-based self-evolving framework that computes answers with fixed programs over a structured parse of each image and checks every fact the programs read, lowering this error to 6\%. The same checks also train the parser without labels. VQS applies across four image domains, and across ten benchmarks it improves Qwen3-VL by up to 3.18 points, beating the strongest baseline at every scale and reaching 3.84 points at 2B after three cycles.

\newpage

\bibliography{iclr2027_conference}
\bibliographystyle{iclr2027_conference}

\appendix

\section{Appendix}

\newcommand{\apptocline}[2]{\noindent\hspace*{#1}\hyperref[#2]{\nameref*{#2}}\dotfill\pageref{#2}\par}
\begingroup
\setlength{\parskip}{2pt}
\noindent\textbf{Contents}\par\vspace{3pt}
\apptocline{0em}{app:hparams}
\apptocline{0em}{app:categories}
\apptocline{0em}{app:results}
\apptocline{1.5em}{app:indepdent-parser}
\apptocline{1.5em}{app:model-families}
\apptocline{0em}{app:prompts}
\apptocline{1.5em}{app:prompts:parse}
\apptocline{1.5em}{app:prompts:para}
\apptocline{1.5em}{app:prompts:blind}
\apptocline{1.5em}{app:prompts:train}
\apptocline{1.5em}{app:prompts:progtriples}
\apptocline{1.5em}{app:prompts:baselines}
\apptocline{0em}{app:examples}
\apptocline{1.5em}{app:traces}
\apptocline{1.5em}{app:executor}
\apptocline{1.5em}{app:vqs_vs_vise}
\endgroup
\vspace{6pt}

\subsection{Training and evaluation settings}
\label{app:hparams}

We train the solver with GRPO~\citep{grpo} on top of a LoRA adapter of rank 64, applied to all linear layers with the vision modules excluded, so the vision tower stays frozen and no visual parameters are trained. We use AdamW at a constant learning rate of $1\mathrm{e}{-5}$, a low-variance KL penalty against a frozen reference with $\beta=0.01$, and clip ratios of $0.2$ and $0.3$ with dual clipping at $3.0$. For each prompt we draw 8 rollouts at temperature $1.0$ with a 256-token response budget and cap training images at $1{,}003{,}520$ pixels. We collect 256 prompts per rollout batch, update on a global batch of 128, and train for 96 steps. The three scales share these settings, except that the per-device micro-batch is 16 at 2B and 4 at 8B. \Cref{tab:hparams} lists the full settings.

\begin{table}[t]
\centering\small
\caption{\textbf{Solver training settings.} Identical across 2B, 4B, and 8B except for the per-device micro-batch.}
\label{tab:hparams}
\begin{tabular}{ll}
\toprule
Setting & Value \\
\midrule
\multicolumn{2}{l}{\textit{Adapter}} \\
LoRA rank, alpha & 64, 128 \\
Target modules & all linear layers, \texttt{.*visual.*} excluded \\
Trainable visual parameters & 0.00M \\
Gradient checkpointing & on \\
\midrule
\multicolumn{2}{l}{\textit{Optimiser}} \\
Algorithm & AdamW, betas $(0.9, 0.999)$, weight decay $0.01$ \\
Learning rate & $1\mathrm{e}{-5}$, constant \\
\midrule
\multicolumn{2}{l}{\textit{GRPO}} \\
Advantage estimator & GRPO \\
KL penalty & low-variance estimator, $\beta = 0.01$, frozen reference \\
Clip ratio (low, high) & $0.2$, $0.3$, dual clipping at $3.0$ \\
PPO epochs per step & 1 \\
Dynamic sampling & off \\
\midrule
\multicolumn{2}{l}{\textit{Rollout}} \\
Rollouts per prompt & 8 at temperature $1.0$, top-$p$ $1.0$ \\
Prompt budget & 2048 tokens \\
Response budget & 256 tokens \\
Training image cap & 1{,}003{,}520 pixels \\
\midrule
\multicolumn{2}{l}{\textit{Batching and schedule}} \\
Rollout batch & 256 prompts \\
Global update batch & 128 \\
Micro-batch per device & 16 at 2B, 4 at 8B \\
GPU memory utilisation & 0.60 \\
Steps & 96 \\
\bottomrule
\end{tabular}
\end{table}

\subsection{Benchmark categories}
\label{app:categories}

We group the ten benchmarks into three categories by the main skill each one tests (\cref{tab:categories}). \textbf{Perception} benchmarks ask about what is visible in the image, such as objects, relations, printed text, and spatial layout. \textbf{Knowledge \& Reasoning} benchmarks also need something the image does not supply, such as outside facts, subject knowledge, or logical reasoning. \textbf{General} benchmarks mix both kinds of questions, so we keep them as a separate group. A category's score is the unweighted mean of its benchmarks, and the overall score is the mean over all ten, as in \cref{tab:main}.

\begin{table}[t]
\centering
\small
\caption{\textbf{Benchmark categories.} Each benchmark belongs to one category.}
\label{tab:categories}
\begin{tabular}{lll}
\toprule
Category & Benchmark & What it tests \\
\midrule
Perception & GQA & objects, attributes, and relations \\
 & InfoVQA & text and layout in infographics \\
 & EmbSpatial (ESB) & spatial relations in embodied scenes \\
\midrule
Knowledge \& Reasoning & OK-VQA & outside knowledge \\
 & ScienceQA (SQA) & school science \\
 & MMMU & college-level subjects \\
 & LogicVista (LogicV) & logical reasoning over images \\
\midrule
General & MMBench (MMB) & many perception and reasoning skills \\
 & MMStar & questions that require the image \\
 & SEED-Bench (SEED) & broad visual understanding \\
\bottomrule
\end{tabular}
\end{table}

\subsection{More Results}\label{app:results}
\begin{table}[t]
\centering
\fontsize{8.5}{10}\selectfont
\setlength{\tabcolsep}{3.5pt}
\caption{\textbf{Checker independence.} All five checkers grade the same 1{,}417 claims. Accepts is the share of the 1{,}170 parser claims a checker calls true. Trap YES is the share of the 247 corrupted claims it wrongly accepts. $\kappa$ is Cohen's kappa with our checker over all claims.}
\rowcolors{2}{verylightgray}{white}
\begin{tabular}{@{}llrrr@{}}
\toprule
\textbf{Checker} & \textbf{Relation to the parser} & \textbf{Accepts (\%)} & \textbf{Trap YES (\%)} & $\boldsymbol{\kappa}$ \textbf{vs.\ ours} \\
\midrule
Qwen3-VL-2B (ours) & same weights & 94.4 & 1.2 & -- \\
Qwen3-VL-8B & same family, 4$\times$ larger & 93.3 & 1.6 & 0.837 \\
InternVL3.5-8B & different family, 8B & 93.1 & 4.0 & 0.810 \\
Qwen2.5-VL-7B & previous Qwen generation & 89.5 & 0.4 & 0.807 \\
SmolVLM2-2.2B & different family, same size & 86.1 & 5.7 & 0.660 \\
\bottomrule
\end{tabular}

\label{tab:checker-family}
\end{table}

\paragraph{Is the checker independent of the parser?}\label{app:indepdent-parser}
The parser, checker, and solver share the same Qwen3-VL weights, so the checker could share the parser's blind spots and confirm its mistakes. To test this, we grade the same claims with five checkers of different families and sizes (\cref{tab:checker-family}). From 150 parses, we extract 1{,}417 atomic claims. Of these, 1{,}170 come from the parser, and 247 are traps in which we change a number or a label, so the correct verdict is always no. All checkers use the same prompt and greedy decoding. The other models only audit our checker and never touch the training data, so training still uses a single model. Our checker rejects 244 of the 247 traps, so it does not simply approve what the parser writes. Agreement between checkers depends more on size than on family. The two 8B checkers from different families agree with each other more ($\kappa{=}0.877$) than our 2B checker agrees with Qwen3-VL-8B ($\kappa{=}0.837$). SmolVLM2-2.2B agrees least with every other checker ($\kappa$ from 0.62 to 0.66) and accepts the most traps (5.7\%), so it is a weaker checker rather than an independent one. An independent check therefore needs a larger model, not just a different family. The risk also depends on the claim type. Among the claims our checker accepts, at least two other checkers reject 26.7\% of diagram arrows, 8.3\% of chart values, and none of the object-existence claims. Diagram arrows and chart values are therefore where the checker most needs improvement. Since another checker's rejection is not ground truth, we treat these rates as warning signs rather than error rates.

\begin{table}[t]
\centering
\caption{\textbf{Results on other model families.} Every model trains on unlabeled images only. Subscripts give the change against that backbone's base. Abbreviations follow \cref{tab:main}. \textit{VQS improves all ten benchmarks for every family}, raising the average by +1.77 on Gemma3-12B-It, +1.51 on InternVL3-8B-Instruct and +1.84 on Llama-3.2-11B-Vision-Instruct.}
\vspace{-1.0em}
\setlength{\tabcolsep}{3pt}
\resizebox{\textwidth}{!}{%
\begin{tabular}{lccccccccccc}
\toprule
\textbf{Method} & \textbf{GQA} & \textbf{OK-VQA} & \textbf{InfoVQA} & \textbf{SQA} & \textbf{MMMU} & \textbf{MMB} & \textbf{ESB} & \textbf{LogicV} & \textbf{MMStar} & \textbf{SEED} & \textbf{Avg.} \\
\midrule
\multicolumn{12}{l}{\textit{Gemma3-12B-It}} \\
\hdashline
Base & 61.20 & 62.40 & 52.10 & 94.30 & 46.50 & 76.40 & 24.50 & 41.80 & 58.90 & 74.20 & 59.23 \\
\rowcolor{rowblue}
\textbf{VQS (ours)} & \textbf{62.02}\up{0.82} & \textbf{62.71}\up{0.31} & \textbf{55.29}\up{3.19} & \textbf{95.23}\up{0.93} & \textbf{50.84}\up{4.34} & \textbf{79.06}\up{2.66} & \textbf{27.70}\up{3.20} & \textbf{42.87}\up{1.07} & \textbf{59.80}\up{0.90} & \textbf{74.47}\up{0.27} & \textbf{61.00}\up{1.77} \\
\midrule
\multicolumn{12}{l}{\textit{InternVL3-8B-Instruct}} \\
\hdashline
Base & 65.40 & 64.10 & 68.80 & 96.50 & 51.20 & 81.30 & 28.10 & 48.60 & 63.70 & 76.90 & 64.46 \\
\rowcolor{rowblue}
\textbf{VQS (ours)} & \textbf{66.01}\up{0.61} & \textbf{64.36}\up{0.26} & \textbf{71.81}\up{3.01} & \textbf{96.71}\up{0.21} & \textbf{54.87}\up{3.67} & \textbf{83.17}\up{1.87} & \textbf{31.99}\up{3.89} & \textbf{49.42}\up{0.82} & \textbf{64.32}\up{0.62} & \textbf{77.08}\up{0.18} & \textbf{65.97}\up{1.51} \\
\midrule
\multicolumn{12}{l}{\textit{Llama-3.2-11B-Vision-Instruct}} \\
\hdashline
Base & 56.00 & 55.40 & 46.90 & 91.80 & 50.70 & 72.80 & 21.20 & 36.50 & 55.10 & 71.30 & 55.77 \\
\rowcolor{rowblue}
\textbf{VQS (ours)} & \textbf{57.13}\up{1.13} & \textbf{56.05}\up{0.65} & \textbf{49.87}\up{2.97} & \textbf{93.47}\up{1.67} & \textbf{53.64}\up{2.94} & \textbf{76.04}\up{3.24} & \textbf{24.18}\up{2.98} & \textbf{37.79}\up{1.29} & \textbf{56.22}\up{1.12} & \textbf{71.68}\up{0.38} & \textbf{57.61}\up{1.84} \\
\bottomrule
\end{tabular}
}
\label{tab:other_families}
\end{table}

\paragraph{Does VQS transfer to other model families?}\label{app:model-families}
We also apply VQS to Gemma3-12B-It~\citep{gemma3}, InternVL3-8B-Instruct~\citep{internvl3}, and Llama-3.2-11B-Vision-Instruct~\citep{llama32} (\cref{tab:other_families}). VQS improves all ten benchmarks for every family, raising the average by 1.77 points on Gemma, 1.51 on InternVL3, and 1.84 on Llama. Gains are large on MMMU, InfoVQA, and EmbSpatial, about 3 to 4 points per family, and smallest on SEEDBench and OK-VQA, under 0.7 points. The weakest base, Llama (55.77), gains the most, and the strongest, InternVL3 (64.46), gains the least. All three gains are smaller than on Qwen3-VL (+2.32 to +3.18, \cref{tab:main}).

\ProvideTextCommandDefault{\textquotedbl}{\char34}

\definecolor{qxPhoto}{HTML}{2E7D5B}
\definecolor{qxChart}{HTML}{2F5D9E}
\definecolor{qxInfo}{HTML}{B5651D}
\definecolor{qxDiag}{HTML}{6F4E8C}
\definecolor{qxMuted}{HTML}{5F6670}
\definecolor{qxSlot}{HTML}{9C4221}
\definecolor{qpPipe}{HTML}{34495E}
\definecolor{qpTrain}{HTML}{1F6F8B}
\definecolor{qpAblate}{HTML}{5B5B66}
\definecolor{qpBase}{HTML}{8E4B2E}

\newcommand{\qpf}[1]{{\color{qxSlot}\ttfamily\char123 #1\char125}}
\newcommand{\qpob}{{\color{qxMuted}\ttfamily\char123}}
\newcommand{\qpcb}{{\color{qxMuted}\ttfamily\char125}}

\lstdefinestyle{qxprompt}{basicstyle=\ttfamily\footnotesize, breaklines=true,
  breakatwhitespace=true, breakindent=0pt, columns=fullflexible, keepspaces=true, upquote=true,
  showstringspaces=false, aboveskip=0pt, belowskip=0pt, moredelim=[s][\color{qxSlot}]{\{}{\}}}

\newtcblisting{qpbox}[2][qpPipe]{enhanced, listing only, breakable, lines before break=5, colback=#1!4!white,
  colframe=#1!80!black, boxrule=0.5pt, arc=2pt, colbacktitle=#1!85!black, coltitle=white,
  fonttitle=\sffamily\bfseries\footnotesize, title={#2}, left=2.5mm, right=2.5mm, top=1.2mm,
  bottom=1.6mm, toptitle=0.7mm, bottomtitle=0.7mm, before skip=5pt, after skip=9pt,
  listing options={style=qxprompt}}

\lstdefinestyle{qxjson}{basicstyle=\ttfamily\scriptsize, columns=fullflexible, keepspaces=true,
  upquote=true, showstringspaces=false, aboveskip=0pt, belowskip=0pt, breaklines=true,
  morestring=[b]", stringstyle=\color{qxSlot}, morekeywords={true,false,null},
  keywordstyle=\bfseries}

\newtcblisting{qpschema}[2][qpPipe]{enhanced, listing only, breakable, lines before break=6,
  colback=#1!3!white, colframe=#1!80!black, boxrule=0.5pt, arc=2pt, colbacktitle=#1!85!black,
  coltitle=white, fonttitle=\sffamily\bfseries\footnotesize, title={#2}, left=2.5mm, right=2.5mm,
  top=1.2mm, bottom=1.6mm, toptitle=0.7mm, bottomtitle=0.7mm, before skip=5pt, after skip=9pt,
  listing options={style=qxjson}}

\newtcolorbox{qpgroup}[2][qpPipe]{enhanced, colback=#1!4!white, colframe=#1!80!black,
  boxrule=0.5pt, arc=2pt, colbacktitle=#1!85!black, coltitle=white,
  fonttitle=\sffamily\bfseries\footnotesize, title={#2}, left=2.5mm, right=2.5mm, top=1mm,
  bottom=1.2mm, toptitle=0.7mm, bottomtitle=0.7mm, before skip=5pt, after skip=9pt}
\newenvironment{qprows}
  {\renewcommand{\arraystretch}{1.3}\setlength{\tabcolsep}{0pt}%
   \begin{tabular}{@{}>{\leavevmode\color{qxMuted}\sffamily\scriptsize\raggedright\arraybackslash}p{2.45cm}@{\hspace{3mm}}%
      >{\leavevmode\ttfamily\footnotesize\raggedright\arraybackslash}p{\dimexpr\linewidth-2.45cm-3mm\relax}@{}}}
  {\end{tabular}}
\newcommand{\qpnote}[1]{\par\smallskip{\color{qxMuted}\sffamily\scriptsize #1}}

\raggedbottom
\subsection{Prompts}
\label{app:prompts}

Text in braces, such as \qpf{q}, is a Python format field filled at call time. The prompts in \cref{app:prompts:baselines} belong to the two baselines we compare against, which are the only places a model is asked to judge or vote on an answer.

\subsubsection{Parsing an image into a structure}
\label{app:prompts:parse}

One prompt per domain, each paired with a JSON schema enforced by constrained decoding, so valid
JSON holds by construction rather than by instruction.

\begin{qpbox}[qxChart]{Charts}
Read this chart. List every series and every data point you can see. Copy `printed_value` exactly as printed on the chart (keep %, $, B, commas). Put the plain number in `numeric_value`. Use the real category and axis names.
\end{qpbox}

\begin{qpbox}[qxDiag]{Diagrams}
Read this diagram. List every labelled part as a node, copying labels verbatim. Set has_arrows to true ONLY if the diagram really draws arrows from one labelled part to another, like a cycle or a food web. Many diagrams only label parts and have no such arrows -- for those set has_arrows to false and leave edges empty. Never invent an arrow.
\end{qpbox}

\begin{qpbox}[qxPhoto]{Natural images}
Describe this photo as structure. List the main objects with their visible attributes and boxes in 0-1000 coordinates as [x0,y0,x1,y1]. Then list how the objects relate to each other, using simple predicates like "on", "holding".
\end{qpbox}

\begin{qpbox}[qxInfo]{Infographics}
Read this infographic. List every labelled figure or statistic as an entry. Copy `printed_value` exactly as printed (keep %, $, commas, K/M/B). Put the plain number in `numeric_value` and the unit in `unit`.
\end{qpbox}

The schemas below are the ones the decoder enforces through vLLM's guided decoding, as JSON Schema
with whitespace compacted. Every field is required and no object admits extra fields, so a parse
can neither omit a field nor add one. A \texttt{numeric\_value} may be \texttt{null}: a value that
is drawn but not legible is recorded as null rather than guessed, and the templates skip it.

\begin{qpschema}[qxChart]{Charts: output schema}
{
  "type": "object",
  "additionalProperties": false,
  "required": ["kind", "title", "x_axis_title", "y_axis_title", "series"],
  "properties": {
    "kind": {"const": "chart"},
    "title": {"type": "string"},
    "x_axis_title": {"type": "string"},
    "y_axis_title": {"type": "string"},
    "series": {
      "type": "array",
      "maxItems": 6,
      "items": {
        "type": "object",
        "additionalProperties": false,
        "required": ["series_name", "points"],
        "properties": {
          "series_name": {"type": "string"},
          "points": {
            "type": "array",
            "maxItems": 24,
            "items": {
              "type": "object",
              "additionalProperties": false,
              "required": ["category", "printed_value", "numeric_value"],
              "properties": {
                "category": {"type": "string"},
                "printed_value": {"type": "string"},
                "numeric_value": {"type": ["number", "null"]}}}}}}}}}
\end{qpschema}

\begin{qpschema}[qxDiag]{Diagrams: output schema}
{
  "type": "object",
  "additionalProperties": false,
  "required": ["kind", "title", "has_arrows", "nodes", "edges"],
  "properties": {
    "kind": {"const": "diagram"},
    "title": {"type": "string"},
    "has_arrows": {"type": "boolean"},
    "nodes": {
      "type": "array",
      "maxItems": 20,
      "items": {
        "type": "object",
        "additionalProperties": false,
        "required": ["node_id", "node_label"],
        "properties": {
          "node_id": {"type": "integer"},
          "node_label": {"type": "string"}}}},
    "edges": {
      "type": "array",
      "maxItems": 30,
      "items": {
        "type": "object",
        "additionalProperties": false,
        "required": ["from_node_id", "to_node_id", "edge_label"],
        "properties": {
          "from_node_id": {"type": "integer"},
          "to_node_id": {"type": "integer"},
          "edge_label": {"type": "string"}}}}}}
\end{qpschema}

\begin{qpschema}[qxPhoto]{Natural images: output schema}
{
  "type": "object",
  "additionalProperties": false,
  "required": ["kind", "objects", "relations"],
  "properties": {
    "kind": {"const": "photo"},
    "objects": {
      "type": "array",
      "maxItems": 10,
      "items": {
        "type": "object",
        "additionalProperties": false,
        "required": ["object_id", "object_name", "attributes", "box_0_1000"],
        "properties": {
          "object_id": {"type": "integer"},
          "object_name": {"type": "string"},
          "attributes": {
            "type": "array",
            "maxItems": 3,
            "items": {"type": "string"}},
          "box_0_1000": {
            "type": "array",
            "minItems": 4,
            "maxItems": 4,
            "items": {"type": "integer"}}}}},
    "relations": {
      "type": "array",
      "maxItems": 12,
      "items": {
        "type": "object",
        "additionalProperties": false,
        "required": ["subject_id", "predicate", "object_id"],
        "properties": {
          "subject_id": {"type": "integer"},
          "predicate": {"type": "string"},
          "object_id": {"type": "integer"}}}}}}
\end{qpschema}

\begin{qpschema}[qxInfo]{Infographics: output schema}
{
  "type": "object",
  "additionalProperties": false,
  "required": ["kind", "title", "entries"],
  "properties": {
    "kind": {"const": "infographic"},
    "title": {"type": "string"},
    "entries": {
      "type": "array",
      "maxItems": 20,
      "items": {
        "type": "object",
        "additionalProperties": false,
        "required": ["entry_label", "printed_value", "numeric_value", "unit"],
        "properties": {
          "entry_label": {"type": "string"},
          "printed_value": {"type": "string"},
          "numeric_value": {"type": ["number", "null"]},
          "unit": {"type": "string"}}}}}}
\end{qpschema}

\subsubsection{Rewriting the templated question}
\label{app:prompts:para}

The template program emits the question; the model then rewrites it. The instruction forbids
adding or removing conditions and restricts the rewrite to words already present, because an
unconstrained rewrite silently changes the computation. The gold answer is never shown and never
touched.

\begin{qpbox}{Paraphrase}
Rewrite this question so it sounds natural, keeping its meaning exactly the same.
Do not answer it. Do not add or remove any condition. Use only the words and entities already in the question. Reply with only the rewritten question.

Question: {q}
\end{qpbox}

\subsubsection{The blind gate}
\label{app:prompts:blind}

The shipped question is answered $k{=}4$ times by the same model \emph{with the image withheld}. 

\begin{qpbox}{Blind gate (no image)}
Answer this question with your best guess. You cannot see the image, so guess from the wording alone. Reply with the answer only, no explanation.

Question: {q}
\end{qpbox}

\subsubsection{Training-time prompts}
\label{app:prompts:train}

Each training item is the image, the question, and a suffix fixing the answer format. The format
string is chosen by the program's answer type.

\begin{qpgroup}[qpTrain]{Answer-format suffix}
\begin{qprows}
answer type = word & Answer the question using a single word or phrase. \\
answer type = number & Answer the question using a single number. \\
multiple-choice pool & Answer with the option's letter from the given choices directly. \\
\end{qprows}
\end{qpgroup}

\subsubsection{Model-written programs (ablation)}
\label{app:prompts:progtriples}

In the ablation where the model writes the question, the answer \emph{and} the program, the parse is supplied as JSON and only the executor decides correctness.

\begin{qpbox}[qpAblate]{Question, answer and program written by the model}
You are given an image and a JSON description of it that was read off the image.

{graph}

Write ONE question about the image, its short answer, and a Python program that COMPUTES that answer from the JSON.

Rules for the program:
- exactly one function: def solve(g): ... where g is the JSON above, as a dict
- return the answer; it will be compared to your "answer" field as a string
- read the values out of g. Do NOT write the answer as a literal in the program
- plain Python only: indexing, loops, if, len, sum, min, max, sorted, round, abs, str, int, float
- no imports; use g["key"] and for x in g["list"]; string methods like .lower() and .strip() are fine

Rules for the question:
- answerable from the image alone by someone who cannot see the JSON
- the answer must be a single word, number or short phrase
- do not name the answer in the question

Return JSON with keys question, answer, program.
\end{qpbox}

\subsubsection{Baseline prompts}
\label{app:prompts:baselines}

These are the prompts of the two baselines we compare our generator against. Both start from the same free-form question prompt. 

\begin{qpbox}[qpBase]{Question generation (both baselines)}
Write one question about this image that can be answered with a single word, phrase or number by looking at the image. Reply with only the question.
\end{qpbox}

\paragraph{Majority vote.} The question is answered $5$ times at $T{=}1.0$ with the standard format suffix, and the plurality answer is kept as the label. There is no prompt beyond the suffix.

\begin{qpbox}[qpBase]{Judge}
You are grading one answer to a question about an image.
Question: {q}
Proposed answer: {a}

Look at the image and decide whether the proposed answer is correct.
Reply with exactly one word: YES if it is correct, NO if it is not.
\end{qpbox}

\ProvideTextCommandDefault{\textquotedbl}{\char34}

\definecolor{qxPhoto}{HTML}{2E7D5B}
\definecolor{qxChart}{HTML}{2F5D9E}
\definecolor{qxInfo}{HTML}{B5651D}
\definecolor{qxDiag}{HTML}{6F4E8C}
\definecolor{qxKeep}{HTML}{2E7D32}
\definecolor{qxDrop}{HTML}{C0392B}
\definecolor{qxMuted}{HTML}{5F6670}
\definecolor{qxSlot}{HTML}{9C4221}
\definecolor{qxCode}{HTML}{F6F7F9}

\newcommand{\qxslot}[1]{{\color{qxSlot}$\langle$\textit{#1}$\rangle$}}
\newcommand{\qxopt}[1]{{\color{qxMuted}[}#1{\color{qxMuted}]}}
\newcommand{\qxok}{{\color{qxKeep}\ding{51}}}
\newcommand{\qxno}{{\color{qxDrop}\ding{55}}}
\newcommand{\qxdot}{\unskip\ {\color{qxMuted}\textperiodcentered}\ \ignorespaces}
\newcommand{\qxsep}{\unskip\ \ {\color{white!70}\textperiodcentered}\ \ \ignorespaces}
\newcommand{\qxnone}[1]{{\color{qxMuted}\itshape #1}}
\newcommand{\qxhits}[2]{\quad\mbox{#1 match #2}}
\newcommand{\qxbadge}[2]{\tcbox[on line, colback=white, colframe=white, coltext=#1, boxrule=0pt,
  arc=1.2pt, boxsep=0pt, left=3pt, right=3pt, top=1.2pt, bottom=1.2pt,
  fontupper=\sffamily\bfseries\fontsize{6}{7}\selectfont]{#2}}
\newcommand{\qxkept}{\qxbadge{qxKeep}{KEPT}}
\newcommand{\qxdropped}[1]{\qxbadge{qxDrop}{DROPPED BY #1}}

\newtcolorbox{qxcard}[3]{enhanced, sidebyside, sidebyside align=top seam, lefthand width=3.6cm,
  sidebyside gap=3.5mm, colback=white, colframe=#1, colbacktitle=#1, coltitle=white,
  boxrule=0.5pt, arc=2pt, left=2mm, right=2mm, top=1.6mm, bottom=1.6mm, toptitle=0.8mm,
  bottomtitle=0.8mm, fonttitle=\sffamily\footnotesize, title={#3},
  before upper={\centering\includegraphics[width=\linewidth,height=3.9cm,keepaspectratio]{#2}},
  before skip=4pt, after skip=6pt}

\newenvironment{qxfields}
  {\scriptsize\renewcommand{\arraystretch}{1.25}\setlength{\tabcolsep}{0pt}%
   \begin{tabular}{@{}>{\leavevmode\color{qxMuted}\sffamily\bfseries\raggedright\arraybackslash}p{1.45cm}%
      >{\leavevmode\raggedright\arraybackslash}p{\dimexpr\linewidth-1.45cm\relax}@{}}}
  {\end{tabular}}

\newtcblisting{qxcode}[1]{enhanced, listing only, colback=qxCode, colframe=black!18, boxrule=0.4pt,
  arc=2pt, left=2mm, right=2mm, top=0.5mm, bottom=0.5mm, title={#1}, colbacktitle=black!6,
  coltitle=black!80, fonttitle=\sffamily\footnotesize,
  listing options={language=Python, basicstyle=\ttfamily\scriptsize, columns=fullflexible,
    keepspaces=true, upquote=true, keywordstyle=\color{qxChart}\bfseries,
    commentstyle=\color{qxMuted}\itshape, stringstyle=\color{qxSlot}, showstringspaces=false}}

\clearpage
\raggedbottom
\subsection{Qualitative examples}
\label{app:examples}

This section shows what the generator produces. Everything in it comes from Qwen3-VL-2B parsing,
checking and gating its own images. \Cref{tab:template_library} lists
twenty template families with a question each one wrote. \Cref{app:traces} follows single
questions from the parse to the answer key, including two that the filters removed, and
\cref{app:executor} shows the code that computes one family's answer. Levels follow the
generator's operand depth: L1 lookup, L2 single hop, L3 two operands, L4 derived operand, L5
comparison of aggregates.

\begin{table}[!h]
\centering
\scriptsize
\caption[Samples of template families, with one question each]{\textbf{Template families, with one question
each.} A slot in \qxslot{angle brackets} is filled from the parse, and a part in
\qxopt{square brackets} appears only when the chart has several series. Each example is a question
the family wrote for a real image, shown after paraphrasing where the paraphrase passed the guard,
with the key its program computed. We checked every key against its image.}
\label{tab:template_library}
\setlength{\tabcolsep}{4pt}
\renewcommand{\arraystretch}{1.25}
\begin{tabular}{@{}>{\raggedright\arraybackslash}p{2.75cm}
                   >{\centering\arraybackslash}p{0.4cm}
                   >{\raggedright\arraybackslash}p{\dimexpr\textwidth-5.5cm-6\tabcolsep\relax}
                   >{\raggedright\arraybackslash}p{2.35cm}@{}}
\toprule
\textbf{Family} & \textbf{Lv} & \textbf{Template} {\color{qxMuted}$\rightarrow$ example question}
  & \textbf{Answer key} \\
\midrule
\multicolumn{4}{@{}l}{\cellcolor{qxPhoto!9}\color{qxPhoto}\sffamily\bfseries Photos} \\
\texttt{existence} & L2 & Are there any \qxslot{objects} in the image? \newline
  {\color{qxMuted}$\rightarrow$ Are there any windows in the image?} & no \\
\texttt{relation\_\allowbreak existence} & L2 & Are there any \qxslot{objects} \qxslot{relation}
  the \qxslot{anchor}? \newline {\color{qxMuted}$\rightarrow$ Are there any chairs to the right of
  the painting?} & yes \\
\texttt{attribute\_\allowbreak filtered\_\allowbreak existence} & L3 & Is there a
  \qxslot{attribute} \qxslot{object} in the picture? \newline {\color{qxMuted}$\rightarrow$ Is
  there white grass in the picture?} & no \\
\texttt{state\_\allowbreak filtered\_\allowbreak existence} & L3 & Do you see a \qxslot{state}
  \qxslot{object} in the picture? \newline {\color{qxMuted}$\rightarrow$ Do you see an open oven
  door in the picture?} & yes \\
\texttt{gated\_\allowbreak object\_\allowbreak recognition} & L3 & Which kind of
  \qxslot{category} is \qxslot{attribute}? \newline {\color{qxMuted}$\rightarrow$ Which kind of
  furniture is green?} & bench \\
\texttt{relation\_\allowbreak chain} & L4 & Are there any \qxslot{objects} \qxslot{relation} the
  \qxslot{anchor} that is \qxslot{relation} the \qxslot{target}? \newline
  {\color{qxMuted}$\rightarrow$ Are there any watches to the left of the man that is to the right
  of the baseball bat?} & yes \\
\multicolumn{4}{@{}l}{\cellcolor{qxChart!9}\color{qxChart}\sffamily\bfseries Charts} \\
\texttt{retrieve} & L1 & What is the value of \qxslot{category}\qxopt{ for \qxslot{series}}?
  \newline {\color{qxMuted}$\rightarrow$ What is the value of Indiana?} & 0.53 \\
\texttt{extremum} & L2 & Which category has the \qxslot{highest\,$|$\,lowest}
  value\qxopt{ for \qxslot{series}}? \newline {\color{qxMuted}$\rightarrow$ Which category has
  the highest value?} & 2025* \\
\texttt{compare} & L2 & Which has a higher value, \qxslot{a} or \qxslot{b}\qxopt{ for
  \qxslot{series}}? \newline {\color{qxMuted}$\rightarrow$ Which has a higher value, 2018 or
  2020?} & 2018 \\
\texttt{diff} & L3 & How much higher is \qxslot{a} than \qxslot{b}\qxopt{ for \qxslot{series}}?
  \newline {\color{qxMuted}$\rightarrow$ How much higher is 2019 than 2018 for Stayovers?}
  & 31984 \\
\texttt{ratio} & L3 & What is the ratio of \qxslot{a} to \qxslot{b}\qxopt{ for
  \qxslot{series}}? \newline {\color{qxMuted}$\rightarrow$ What is the ratio of Somewhat prefer
  theater to Strongly prefer streaming for June 2020?} & 0.44 \\
\texttt{series\_\allowbreak compare} & L3 & In \qxslot{category}, which is higher,
  \qxslot{series} or \qxslot{series}? \newline {\color{qxMuted}$\rightarrow$ In 2016, which is
  higher, China or United States?} & China \\
\texttt{ranked\_\allowbreak arithmetic} & L5 & Is the sum of the two highest values greater than
  the sum of the three lowest? & No \\
\multicolumn{4}{@{}l}{\cellcolor{qxInfo!9}\color{qxInfo}\sffamily\bfseries Infographics} \\
\texttt{ig\_\allowbreak chart\_\allowbreak metadata} & L1 & What is the title of this chart?
  & THE WORLD'S 10 LARGEST STADIUMS \\
\texttt{top\_\allowbreak k} & L2 & Which are the \qxslot{k} highest\qxopt{ in \qxslot{title}}?
  \newline {\color{qxMuted}$\rightarrow$ Which are the 3 highest in HOW AMERICANS RATE BUSINESS
  FIGURES?} & Bill Gates, Warren Buffett and Mark Cuban \\
\texttt{multi\_\allowbreak item\_\allowbreak filter} & L2 & Which have a value greater than
  \qxslot{threshold}\qxopt{ in \qxslot{title}}? \newline {\color{qxMuted}$\rightarrow$ Which
  have a value greater than 37 in HOW AMERICANS RATE BUSINESS FIGURES?} & Bill Gates, Warren
  Buffett, Mark Cuban and Elon Musk \\
\texttt{ig\_\allowbreak diff} & L3 & How much higher is \qxslot{a} than \qxslot{b}? \newline
  {\color{qxMuted}$\rightarrow$ How much higher is China than the U.S.?} & 994 \\
\texttt{compound\_\allowbreak lookup} & L3 & Which is highest\qxopt{ in \qxslot{title}} and what
  is its value? \newline {\color{qxMuted}$\rightarrow$ Which is highest in HOW AMERICANS RATE
  BUSINESS FIGURES and what is its value?} & Bill Gates, 45 \\
\multicolumn{4}{@{}l}{\cellcolor{qxDiag!9}\color{qxDiag}\sffamily\bfseries Diagrams} \\
\texttt{diagram\_\allowbreak identification} & L1 & What is shown in this diagram? \newline
  {\color{qxMuted}$\rightarrow$ What is depicted in this diagram?} & Types of Volcanoes \\
\texttt{existence} & L2 & Does every labelled item in this diagram have an arrow attached? & yes \\
\bottomrule
\end{tabular}
\end{table}

\clearpage
\subsubsection{Traced questions}
\label{app:traces}

Each card follows one generated question through the pipeline. \textsf{\bfseries Parse} lists the fields the template read, \textsf{\bfseries Program} lists the operations that compute the answer, and \textsf{\bfseries Question} gives the template question with its paraphrase. \textsf{\bfseries Fact check} gives the checker's yes or no verdict on each claim behind the answer, and \textsf{\bfseries Blind gate} lists four answers from the same model without the image (\cref{eq:band}). The last two cards show questions that the filters removed.

\begin{qxcard}{qxPhoto}{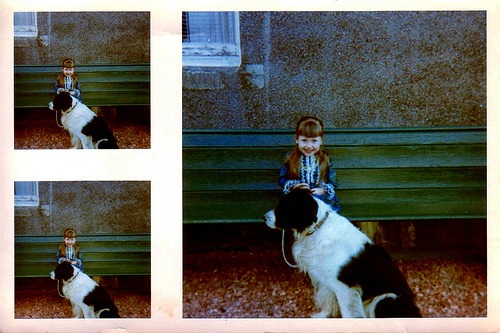}{\textbf{Photo}\qxsep
gated\_object\_recognition\qxsep L3\hfill\qxkept}
\tcblower
\begin{qxfields}
Parse & furniture objects whose color is green: \{bench\} \\
Program & \texttt{gated\_object\_recognition(category=furniture, color=green)}, which must find
  exactly one kind \\
Question & Which kind of furniture is green? \\
Fact check & read back the bench's color: ``green'' \qxok\ \qxnone{(attribute read-back)} \\
Blind gate & sofa\qxdot couch\qxdot ``The green furniture is a chair.''\qxdot ``leafy green
  furniture''\qxhits{0/4}{\qxok} \\
Answer key & \textbf{bench} \\
\end{qxfields}
\end{qxcard}

\begin{qxcard}{qxPhoto}{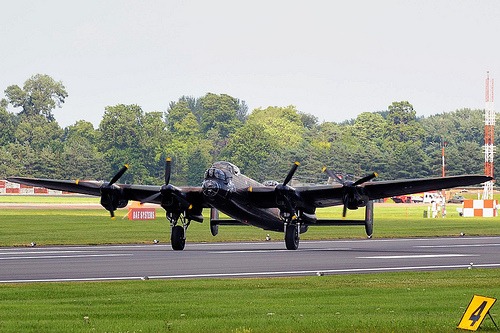}{\textbf{Photo}\qxsep
attribute\_filtered\_existence\qxsep L3\hfill\qxkept}
\tcblower
\begin{qxfields}
Parse & grass objects carrying the attribute white: none \\
Program & \texttt{attribute\_filtered\_existence(object=grass, attribute=white)} \\
Question & Is there a white grass in the picture?\newline
  $\rightarrow$ Is there white grass in the picture? \qxnone{(paraphrase accepted)} \\
Fact check & asked whether there is white grass: ``no'' \qxok \\
Blind gate & Yes\qxdot Yes\qxdot Yes\qxdot Yes\qxhits{0/4}{\qxok} \\
Answer key & \textbf{no} \\
\end{qxfields}
\end{qxcard}

\begin{qxcard}{qxChart}{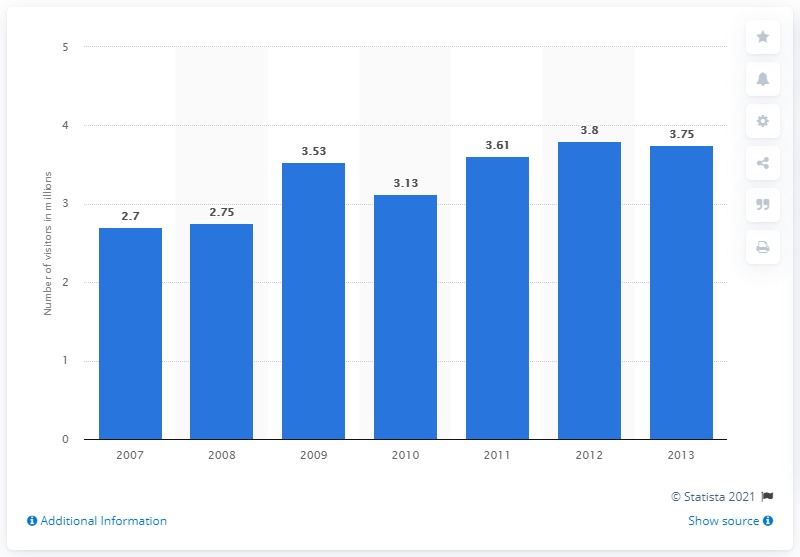}{\textbf{Chart}\qxsep
ranked\_arithmetic\qxsep L5\hfill\qxkept}
\tcblower
\begin{qxfields}
Parse & \textit{Number of visitors in millions}: 2007\,2.7\qxdot 2008\,2.75\qxdot
  2009\,3.53\qxdot 2010\,3.13\qxdot 2011\,3.61\qxdot 2012\,3.8\qxdot 2013\,3.75 \\
Program & two highest $3.8+3.75=7.55$; three lowest $2.7+2.75+3.13=8.58$;
  $7.55>8.58$ is false \\
Question & Is the sum of the two highest values greater than the sum of the three lowest? \\
Fact check & 7/7 values read back exactly, e.g.\ 2009 $\rightarrow$ 3.53 \qxok \\
Blind gate & Yes\qxdot Yes\qxdot Yes\qxdot Yes\qxhits{0/4}{\qxok} \\
Answer key & \textbf{No} \\
\end{qxfields}
\end{qxcard}

\begin{qxcard}{qxChart}{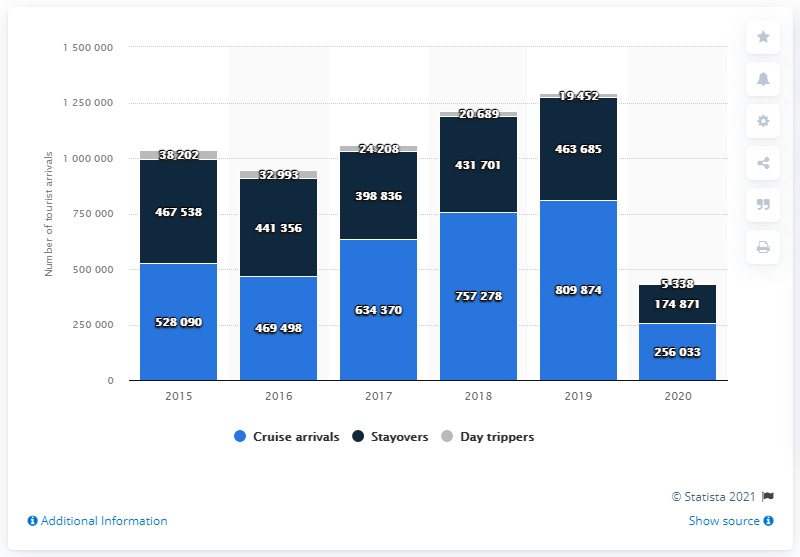}{\textbf{Chart}\qxsep
diff\qxsep L3\hfill\qxkept}
\tcblower
\begin{qxfields}
Program & $|\,\text{Stayovers}(2019)-\text{Stayovers}(2018)\,|$ on a stacked chart with three
  series \\
Question & How much higher is 2019 than 2018 for Stayovers? \\
Fact check & read back 463,685 and 431,701; the difference recomputed from them, 31,984, matches
  the key \qxok \\
Blind gate & 10\qxdot 1\qxdot 10\qxdot 1\qxhits{0/4}{\qxok} \\
Answer key & \textbf{31984} \\
\end{qxfields}
\end{qxcard}

\begin{qxcard}{qxChart}{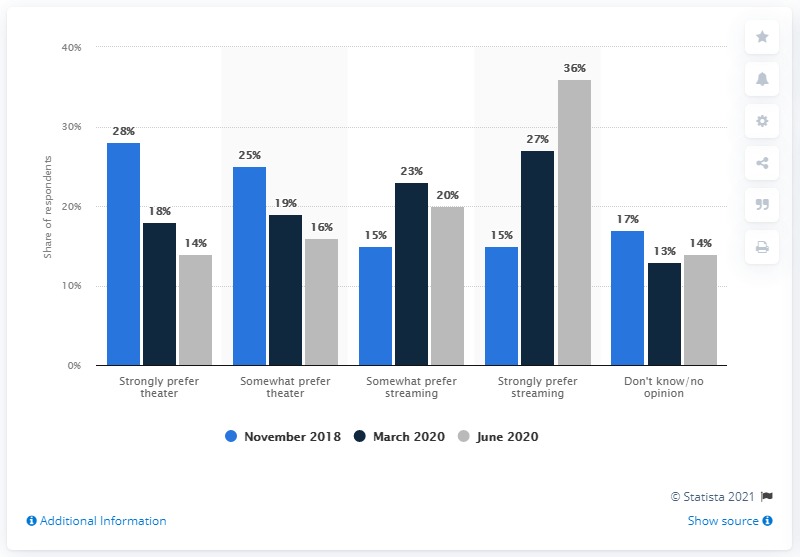}{\textbf{Chart}\qxsep
ratio\qxsep L3\hfill\qxkept}
\tcblower
\begin{qxfields}
Program & Somewhat prefer theater $\div$ Strongly prefer streaming, in the June 2020 series \\
Question & What is the ratio of Somewhat prefer theater to Strongly prefer streaming for June
  2020? \\
Fact check & read back 16 and 36; the ratio recomputed from them, $16/36=0.444$, matches the key
  \qxok \\
Blind gate & 1.2\qxdot 0.22\qxdot 0.67\qxdot 0.50\qxhits{0/4}{\qxok} \\
Answer key & \textbf{0.44} \\
\end{qxfields}
\end{qxcard}

\begin{qxcard}{qxInfo}{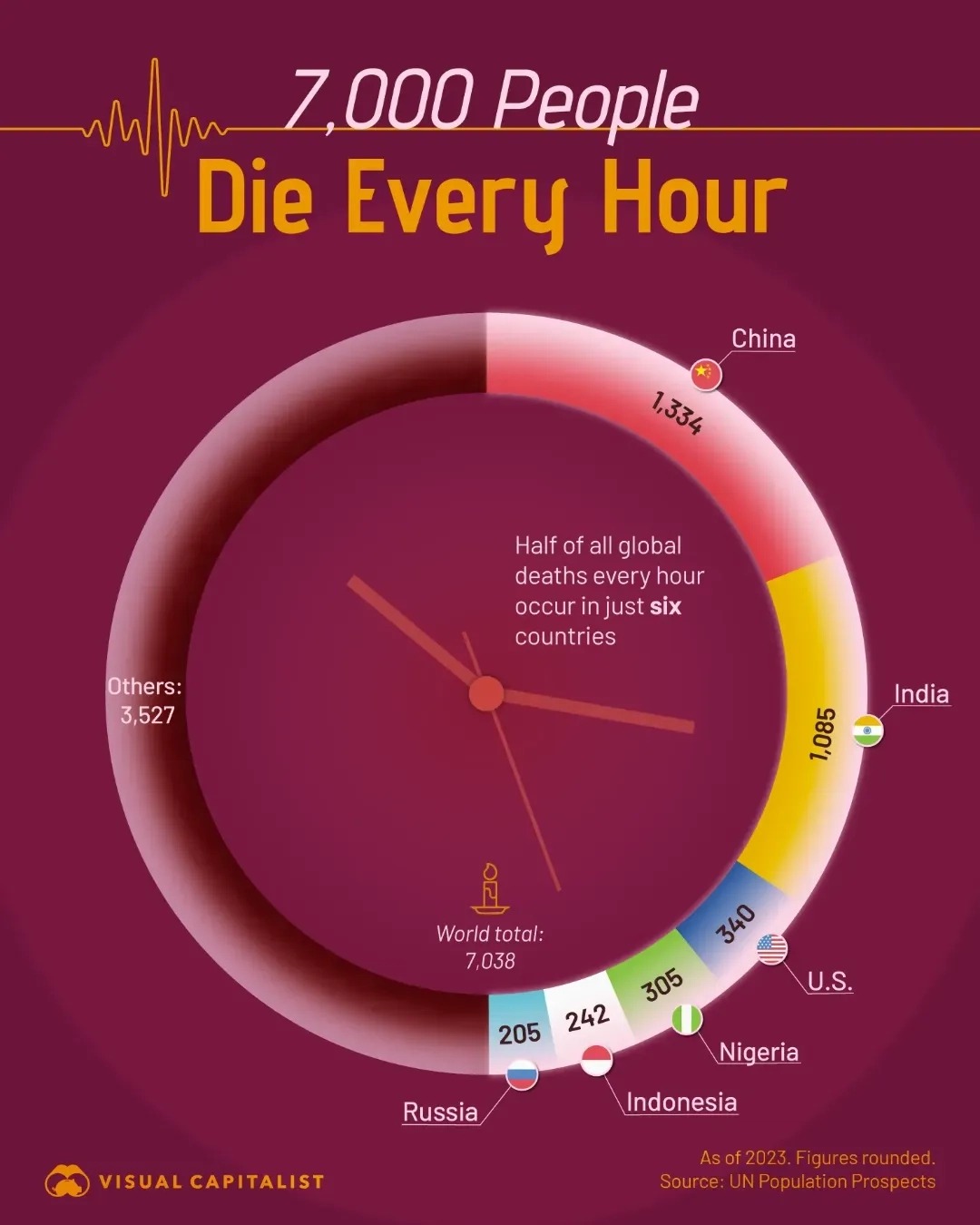}{\textbf{Infographic}\qxsep
ig\_diff\qxsep L3\hfill\qxkept}
\tcblower
\begin{qxfields}
Program & $|\,\text{China}-\text{U.S.}\,|$ \\
Question & How much higher is China than U.S.?\newline
  $\rightarrow$ How much higher is China than the U.S.? \qxnone{(paraphrase accepted)} \\
Fact check & read back 1,334 and 340; the difference recomputed from them, 994, matches the key
  \qxok \\
Blind gate & ``10000000''\qxdot ``China is taller than the U.S.\ by about 3,000 miles.''\qxdot
  two more free-text guesses\qxhits{0/4}{\qxok} \\
Answer key & \textbf{994} \\
\end{qxfields}
\end{qxcard}

\begin{qxcard}{qxInfo}{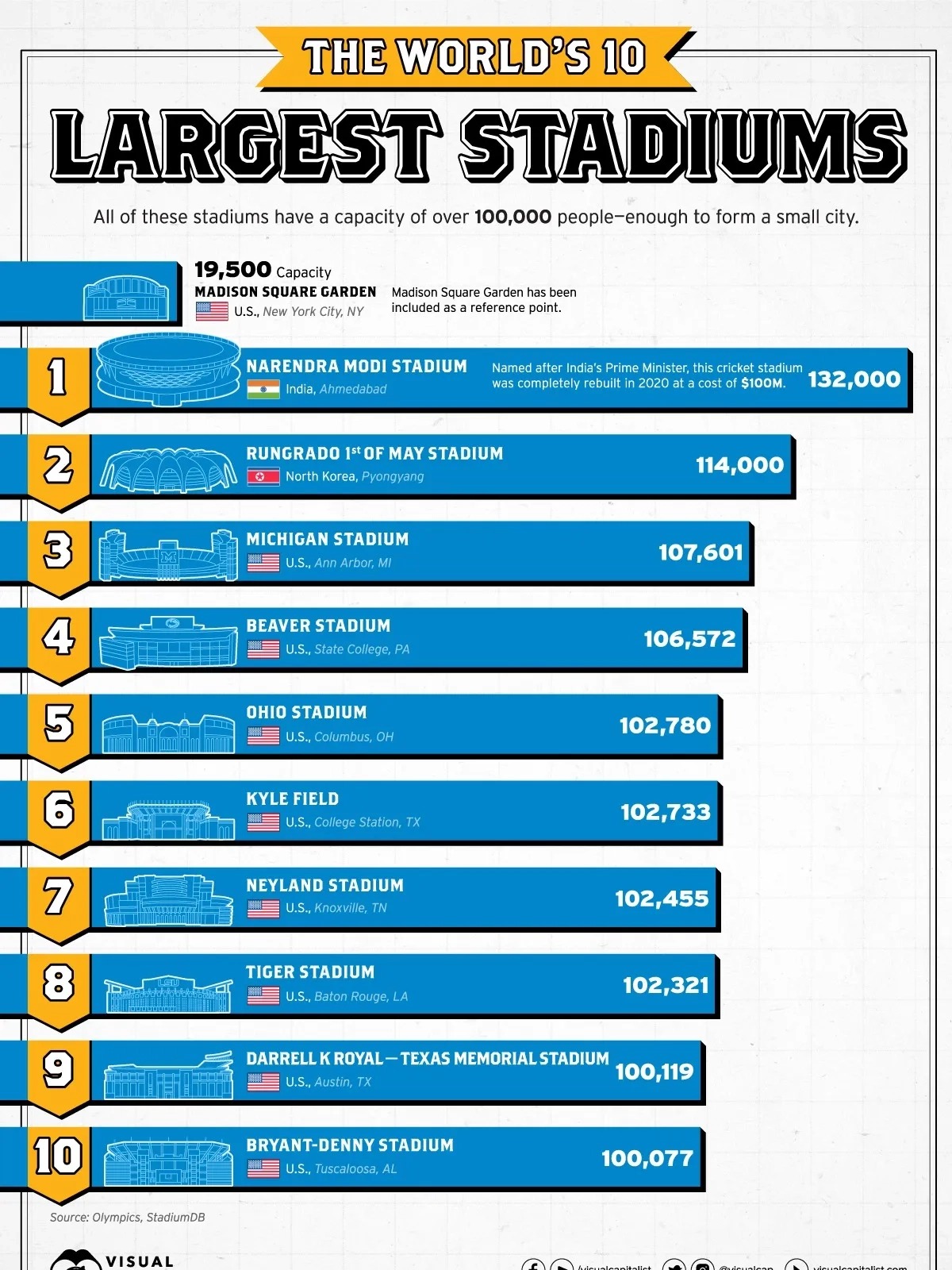}{\textbf{Infographic}\qxsep
ig\_chart\_metadata\qxsep L1\hfill\qxkept}
\tcblower
\begin{qxfields}
Parse & title: THE WORLD'S 10 LARGEST STADIUMS \\
Program & read the title field of the parse \\
Question & What is the title of this chart? \\
Fact check & read back ``The World's 10 Largest Stadiums'' \qxok\ (case is ignored) \\
Blind gate & ``Line Chart''\qxdot ``Average household income by age group''\qxdot two free-text
  guesses\qxhits{0/4}{\qxok} \\
Answer key & \textbf{THE WORLD'S 10 LARGEST STADIUMS} \\
\end{qxfields}
\end{qxcard}

\begin{qxcard}{qxDiag}{}{\textbf{Diagram}\qxsep
diagram\_identification\qxsep L1\hfill\qxkept}
\tcblower
\begin{qxfields}
Parse & title: Types of Volcanoes \\
Program & read the title field of the parse \\
Question & What is shown in this diagram?\newline
  $\rightarrow$ What is depicted in this diagram? \qxnone{(paraphrase accepted)} \\
Fact check & read back the title: ``Types of Volcanoes'' \qxok \\
Blind gate & ``The diagram depicts a chloroplast.''\qxdot ``A bird in the sky.''\qxdot two more
  free-text guesses\qxhits{0/4}{\qxok} \\
Answer key & \textbf{Types of Volcanoes} \\
\end{qxfields}
\end{qxcard}

\begin{qxcard}{qxChart}{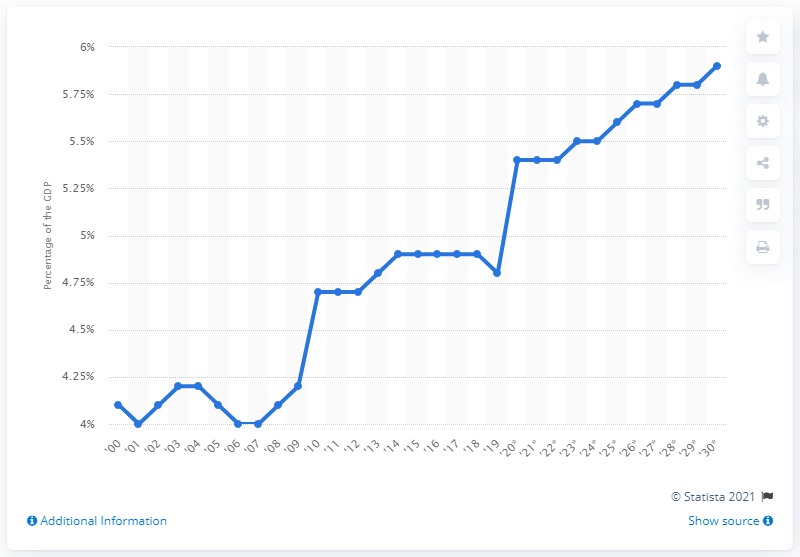}{\textbf{Chart}\qxsep
ranked\_aggregate\qxsep L4\hfill\qxdropped{FACT CHECK}}
\tcblower
\begin{qxfields}
Parse & \textit{Percentage of the GDP}: '00\,4.1\qxdot '01\,3.9\qxdot \dots\qxdot
  '20\,4.7\qxdot '21\,4.8\qxdot '22\,4.8\qxdot '23\,4.8 \qxnone{(24 values; the parse stops at
  '23)} \\
Program & three highest $4.8+4.8+4.8=14.4$ \\
Question & What is the sum of the three highest values? \qxnone{(paraphrase ``three largest''
  rejected: new content word)} \\
Fact check & 21 of 24 read-backs pass; '21 read back as 5.45 \qxno, '22 as 5.4 \qxno, '23 as 5.45
  \qxno. The chart shows 5.4 to 5.5 for these years and reaches 5.9 by '30*, so the key 14.4 was
  wrong. \\
Answer key & \textbf{14.4}, never used \\
\end{qxfields}
\end{qxcard}

\begin{qxcard}{qxChart}{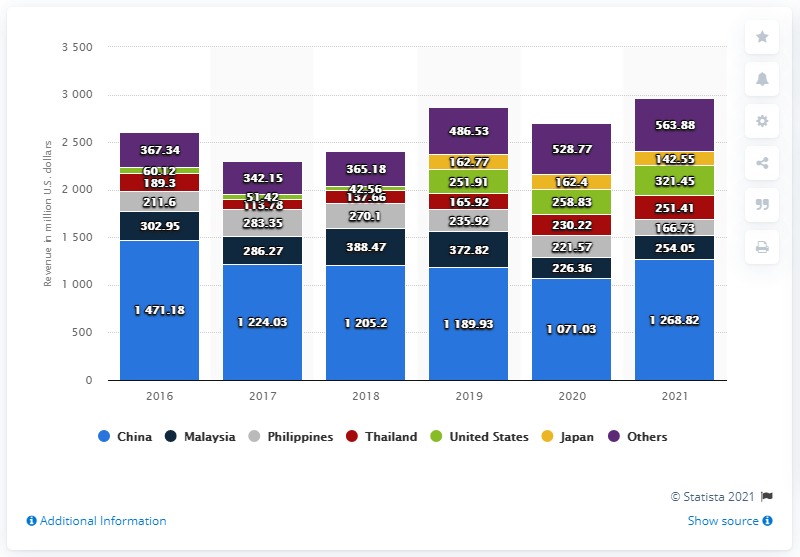}{\textbf{Chart}\qxsep
series\_compare\qxsep L3\hfill\qxdropped{BLIND GATE}}
\tcblower
\begin{qxfields}
Program & compare the China and United States series at 2016, on a stacked chart with seven
  series \\
Question & In 2016, which is higher, China or United States? \\
Fact check & read back 1471.18 for China \qxok\ and 189.3 for United States, where the
  chart shows 60.12 (189.3 is Thailand's value). The comparison recomputed from the two
  read-backs, China, still matches the key \qxok, so the probe passes despite the misread. \\
Blind gate & China\qxdot China\qxdot China\qxdot China\qxhits{4/4}{\qxno}. The key is correct,
  but a model without the image answers it from world knowledge, so the question teaches nothing
  about the image. \\
Answer key & \textbf{China}, never used \\
\end{qxfields}
\end{qxcard}

\subsubsection{The program behind a template}
\label{app:executor}

A family is a question template paired with an executor that computes the answer from the parse.
Below is the pair for \texttt{ranked\_arithmetic}, the family of the third card. The executor
returns nothing rather than a guess when the parse cannot support a unique answer: too few values,
or a tie. The cells it read (\texttt{allprov}) are what the fact check later asks the model to
verify.

\begin{qxcode}{\texttt{ranked\_arithmetic}: template and executor}
# question side: the template, filled from the parse
q = (f"Is the sum of the two highest {subj or 'values'}{sr} greater than the "
     f"sum of the three lowest?")

# answer side: the executor, run on the parsed values of one series
if fam == "ranked_arithmetic":
    k, m = int(prog["k"]), int(prog["m"])        # here k = 2, m = 3
    if len(vals) < k + m:
        return None                              # too few values: no question
    vs = sorted(vals, reverse=True)
    top, bot = sum(vs[:k]), sum(vs[-m:])
    if top == bot:
        return None                              # a tie has no answer: no question
    return ("Yes" if top > bot else "No"), "free", allprov
\end{qxcode}

\definecolor{qxVISE}{HTML}{C8741F}
\definecolor{qxVQS}{HTML}{2E6DB4}
\definecolor{qxBench}{HTML}{3B4252}

\newcommand{\qxours}{{\color{qxVQS}VQS (ours)}}
\newcommand{\qxvise}{{\color{qxVISE}VISE}}
\newcommand{\qxbase}{{\color{qxMuted}Base}}
\newcommand{\qxwhylabel}{{\color{qxDrop}Why wrong}}
\newcommand{\qxright}[1]{#1\ \qxok}
\newcommand{\qxwrong}[1]{{\color{qxDrop}#1}\ \qxno}
\newcommand{\qxbad}[1]{{\color{qxDrop}\bfseries #1}}
\newcommand{\qxgood}[1]{{\color{qxKeep}\bfseries #1}}
\newcommand{\qxbimg}[2][3.9cm]{\centering\includegraphics[width=\linewidth,height=#1,keepaspectratio]{#2}}

\newtcolorbox{qxbench}[2]{enhanced, sidebyside, sidebyside align=top seam, lefthand width=3.6cm,
  sidebyside gap=3.5mm, colback=white, colframe=qxBench, colbacktitle=qxBench, coltitle=white,
  boxrule=0.5pt, arc=2pt, left=2mm, right=2mm, top=1.6mm, bottom=1.6mm, toptitle=0.8mm,
  bottomtitle=0.8mm, fonttitle=\sffamily\footnotesize,
  title={\textbf{#1}\hfill{\color{white!72}sample #2}}, before skip=4pt, after skip=6pt}

\newtcolorbox{qxresponses}[1]{enhanced, colback=white, colframe=qxBench, colbacktitle=qxBench!10,
  coltitle=qxBench, boxrule=0.5pt, arc=2pt, left=2mm, right=2mm, top=1.4mm, bottom=1.4mm,
  toptitle=0.8mm, bottomtitle=0.8mm, fonttitle=\sffamily\bfseries\footnotesize, title={#1},
  before skip=0pt, after skip=8pt}
\newenvironment{qxresprows}
  {\renewcommand{\arraystretch}{1.3}\setlength{\tabcolsep}{0pt}%
   \begin{tabular}{@{}>{\leavevmode\sffamily\bfseries\scriptsize\raggedright\arraybackslash}p{1.45cm}%
      >{\leavevmode\ttfamily\scriptsize\raggedright\arraybackslash}p{\dimexpr\linewidth-1.45cm\relax}@{}}}
  {\end{tabular}}

\subsubsection{VQS, VISE and the base model on the same benchmark questions}
\label{app:vqs_vs_vise}

Each card shows one benchmark item answered by three 2B models: the base model, VISE, and VQS after one training cycle. All responses are taken from our lmms-eval runs, where the three models received the same prompt. \textcolor{qxDrop}{Red text explains why the other two answers are wrong.}

\begin{qxbench}{GQA}{420}
\qxbimg{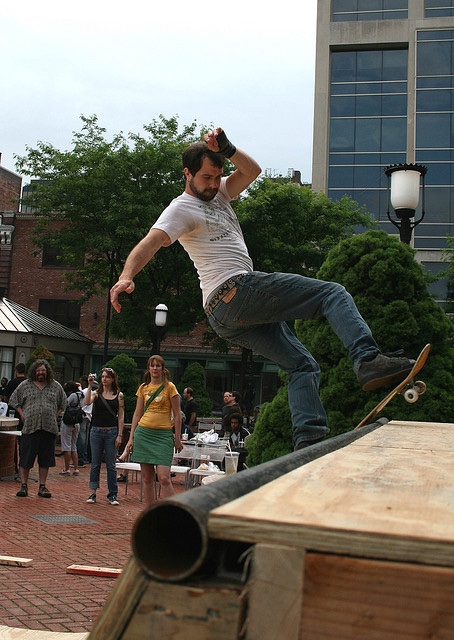}
\tcblower
\begin{qxfields}
Question & Is the woman near the man standing on the bricks? \\
Gold & \textbf{yes} \\
\qxours & \qxright{yes} \\
\qxvise & \qxwrong{no} \\
\qxbase & \qxwrong{no} \\
\qxwhylabel & \color{qxDrop} The woman beside the bearded man in the background is standing on
  the brick pavement. \\
\end{qxfields}
\end{qxbench}

\begin{qxbench}{OK-VQA}{3993}
\qxbimg{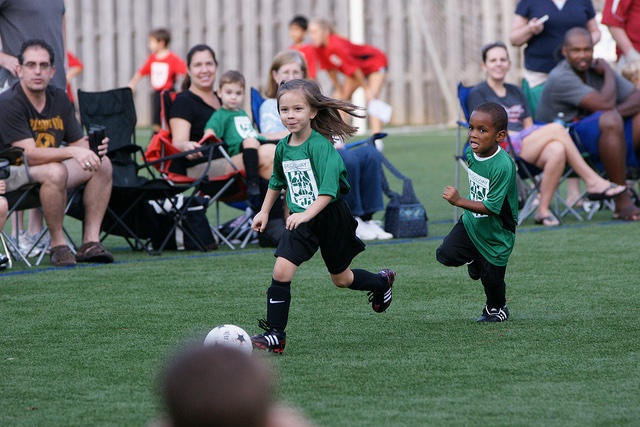}
\tcblower
\begin{qxfields}
Question & How old are the boys? \\
Gold & \textbf{5} \\
\qxours & \qxright{5} \\
\qxvise & \qxwrong{7} \\
\qxbase & \qxwrong{7} \\
\qxwhylabel & \color{qxDrop} None of the ten human annotators answered 7, so it scores 0 under
  VQA accuracy; the players are small children. \\
\end{qxfields}
\end{qxbench}

\begin{qxbench}{InfoVQA}{150}
\qxbimg[3.3cm]{}\par\smallskip
\qxbimg{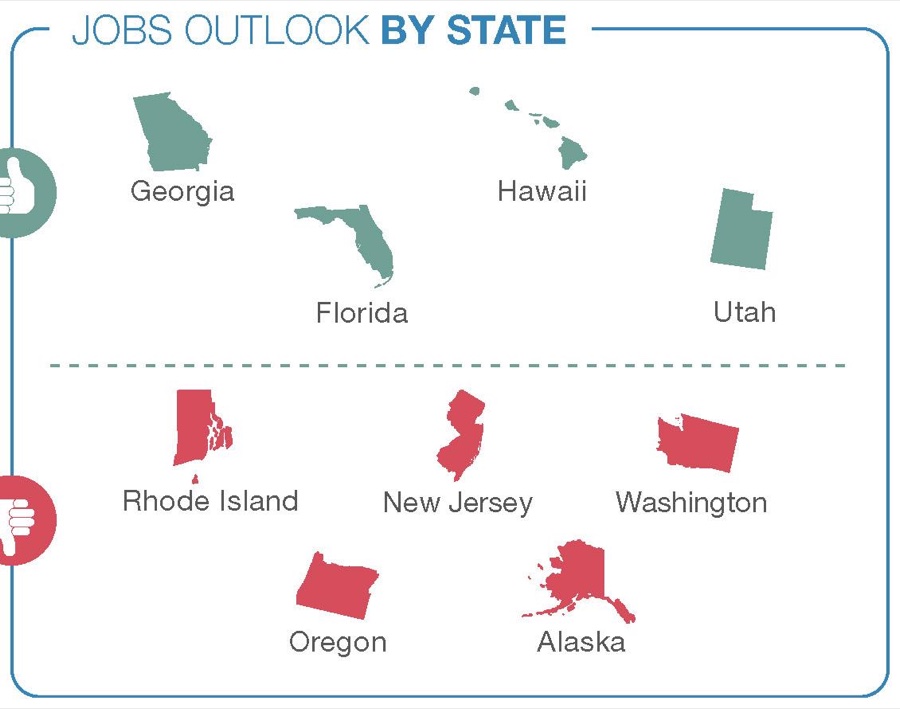}\par
{\centering\color{qxMuted}\sffamily\tiny detail of the ``by state'' panel\par}
\tcblower
\begin{qxfields}
Question & How many states are having a poor job Outlook? \\
Gold & \textbf{5} \\
\qxours & \qxright{5} \\
\qxvise & \qxwrong{4} \\
\qxbase & \qxwrong{4} \\
\qxwhylabel & \color{qxDrop} Five states sit under the thumbs-down in ``Jobs outlook by state'':
  Rhode Island, New Jersey, Washington, Oregon and Alaska. Four is the size of the thumbs-up group
  (Georgia, Hawaii, Florida, Utah). \\
\end{qxfields}
\end{qxbench}

\begin{qxbench}{ScienceQA}{934}
\qxbimg{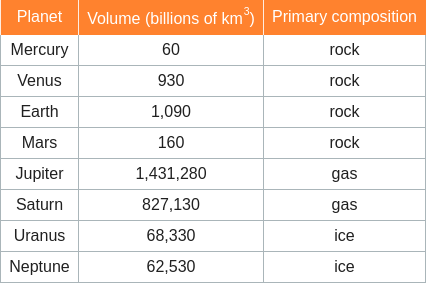}
\tcblower
\begin{qxfields}
Question & Context: Use the data to answer the question below. Is the following statement about
  our solar system true or false? The volume of Mars is more than three times as large as
  Mercury's. A.~false B.~true \\
Gold & \textbf{A} \\
\qxours & \qxright{A. false} \\
\qxvise & \qxwrong{B. true} \\
\qxbase & \qxwrong{B. true} \\
\qxwhylabel & \color{qxDrop} The table gives Mars 160 and Mercury 60 billion km$^3$, so Mars is
  $160/60 \approx 2.7$ times Mercury's volume, less than three. \\
\end{qxfields}
\end{qxbench}

\begin{qxbench}{MMMU}{758}
\qxbimg{}
\tcblower
\begin{qxfields}
Question & Look at the two pictures below. Which picture appears to have more order? A.~a B.~b
  C.~same of the two \\
Gold & \textbf{A} \\
\qxours & \qxright{A. a} \\
\qxvise & \qxwrong{B. b} \\
\qxbase & \qxwrong{B. b} \\
\qxwhylabel & \color{qxDrop} In (b) the jar lies on its side and the marbles are scattered over
  the table and the floor; in (a) they are all inside the upright jar, the more ordered state. \\
\end{qxfields}
\end{qxbench}

\begin{qxbench}{MMBench}{98}
\qxbimg{}
\tcblower
\begin{qxfields}
Question & Select the best answer. Which property do these three objects have in common?
  A.~blue B.~smooth C.~flexible \\
Gold & \textbf{B} \\
\qxours & \qxright{B} \\
\qxvise & \qxwrong{C} \\
\qxbase & \qxwrong{C} \\
\qxwhylabel & \color{qxDrop} Only the silk tie is flexible; the glass bottle and the wooden
  baseball bat are rigid. All three are smooth. \\
\end{qxfields}
\end{qxbench}

\begin{qxbench}{EmbSpatial}{3056}
\qxbimg{}
\tcblower
\begin{qxfields}
Question & Which object, in relation to your current position, holds the farthest placement in
  the image? A) chair B) Unknown C) box D) shelves \\
Gold & \textbf{D} \\
\qxours & \qxright{D} \\
\qxvise & \qxwrong{C} \\
\qxbase & \qxwrong{C} \\
\qxwhylabel & \color{qxDrop} The box is the grey block in the foreground, the object nearest the
  camera. The shelves on the desk at the back of the room are the farthest. \\
\end{qxfields}
\end{qxbench}

\begin{qxbench}{LogicVista}{255}
\qxbimg{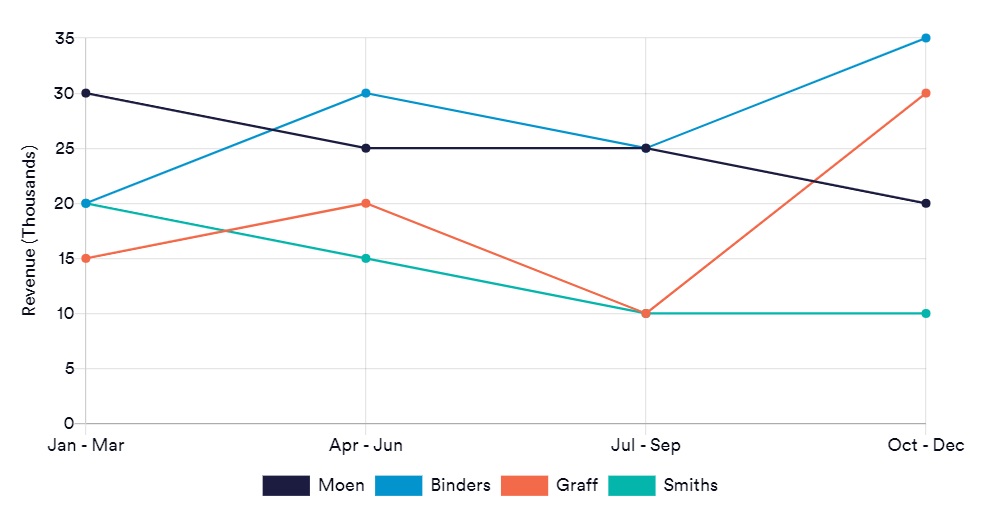}
\tcblower
\begin{qxfields}
Question & Assuming the report has overestimated revenues by 15\%, what was the actual revenue
  generated by Binders in the quarter October to December? Select from A, B, C, and D.
  (A)~30,435 (B)~25,500 (C)~29,750 (D)~35,000 \\
Gold & \textbf{A} \\
\qxours & \qxright{A} \\
\qxvise & \qxwrong{C} \\
\qxbase & \qxwrong{C} \\
\qxwhylabel & \color{qxDrop} A 15\% overestimate means the reported 35 is 1.15 times the actual
  value, so the actual revenue is $35/1.15 \approx 30.4$ thousand. Both baselines take 85\% of
  the reported value instead, and both say the chart covers three quarters when it shows four.
  VISE also ends with a malformed \texttt{\textless CAnswer\textgreater} tag. \\
\end{qxfields}
\end{qxbench}

\begin{qxresponses}{LogicVista, sample 255: full responses}
\begin{qxresprows}
\qxours & ...The chart shows the revenue in thousands for four different companies over four
  quarters. The blue line represents Binders. In the "Oct - Dec" quarter, the chart indicates a
  revenue of 35 thousand. Since the report has overestimated revenues by 15\%, we need to
  calculate the actual revenue by \qxgood{dividing the reported value by (1 + 0.15)}. This gives
  us \qxgood{35 / 1.15 $\approx$ 30.435}. Therefore, the actual revenue generated by Binders in
  the quarter October to December is approximately 30,435.
  \textless answer\textgreater A\textless /answer\textgreater \\[3pt]
\qxvise & ...The chart shows the revenue for four companies (Moen, Binders, Graff, and Smiths)
  \qxbad{over three quarters}. The y-axis represents revenue in thousands of dollars, and the
  x-axis represents the time period.\par\smallskip
  First, we need to determine the reported revenue for Binders in the "Oct - Dec" quarter from
  the chart. The blue line represents Binders. At the "Oct - Dec" point on the x-axis, the
  reported revenue is 35,000 (thousands).\par\smallskip
  The report has overestimated revenues by 15\%. This means that \qxbad{the actual revenue is
  100\% - 15\% = 85\% of the reported value}.\par\smallskip
  So, the actual revenue is:\newline
  \qxbad{(85/100) * 35,000 = 0.85 * 35,000 = 29,750}\par\smallskip
  Therefore, the actual revenue generated by Binders in the quarter October to December was
  29,750.\par\smallskip
  \qxbad{\textless CAnswer\textgreater{} C \textless /CAnswer\textgreater} \\[3pt]
\qxbase & ...The chart shows the revenue for four companies (Moen, Binders, Graff, and Smiths)
  \qxbad{over three quarters}. The y-axis represents revenue in thousands of dollars, and the
  x-axis represents the time period.\par\smallskip
  First, we need to determine the reported revenue for Binders in the "Oct - Dec" quarter from
  the chart. The blue line represents Binders. At the "Oct - Dec" mark on the x-axis, the
  reported revenue is 35 thousand dollars.\par\smallskip
  The report has overestimated revenues by 15\%. This means that \qxbad{the actual revenue is
  100\% - 15\% = 85\% of the reported value}.\par\smallskip
  So, the actual revenue is:\newline
  \qxbad{(85/100) * 35,000 = 0.85 * 35,000 = 29,750}\par\smallskip
  Therefore, the actual revenue generated by Binders in the quarter October to December was
  29,750.\par\smallskip
  \textless answer\textgreater C\textless /answer\textgreater \\
\end{qxresprows}
\end{qxresponses}

\begin{qxbench}{MMStar}{152}
\qxbimg{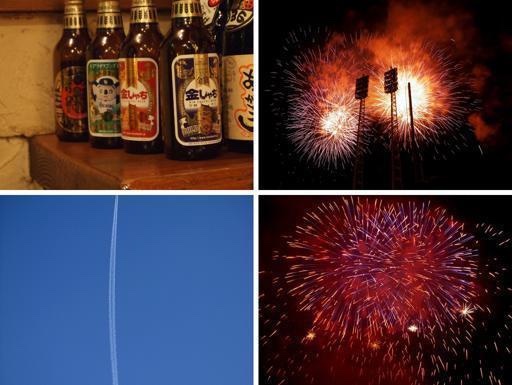}
\tcblower
\begin{qxfields}
Question & Which image is the brightest one? Options: A: upper left, B: upper right, C: down
  left, D: down right \\
Gold & \textbf{C} \\
\qxours & \qxright{C} \\
\qxvise & \qxwrong{B} \\
\qxbase & \qxwrong{B} \\
\qxwhylabel & \color{qxDrop} The upper-right panel is fireworks against a night sky and mostly
  black; the lower-left panel, a clear daytime sky, is bright across the whole frame. \\
\end{qxfields}
\end{qxbench}

\begin{qxbench}{SEED-Bench}{2433}
\qxbimg{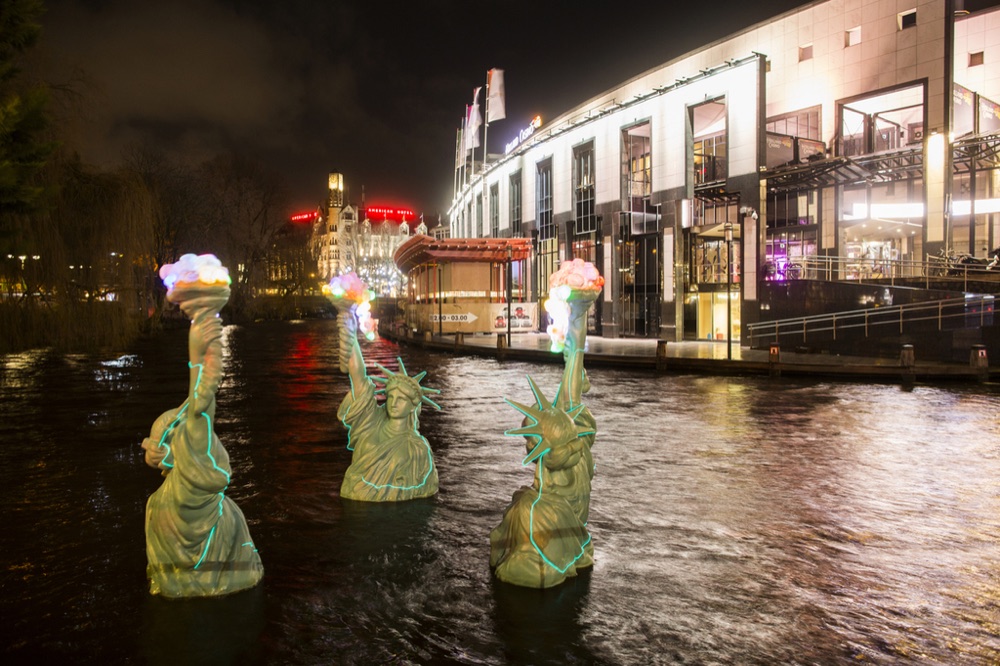}
\tcblower
\begin{qxfields}
Question & Which of the following objects are detected in the image? A.~Person B.~Yellow and blue
  sculpture C.~Ball D.~All of the above \\
Gold & \textbf{B} \\
\qxours & \qxright{B} \\
\qxvise & \qxwrong{D} \\
\qxbase & \qxwrong{D} \\
\qxwhylabel & \color{qxDrop} There is no person and no ball in the image, so ``All of the above''
  cannot hold; the human figures are illuminated Statue of Liberty sculptures. \\
\end{qxfields}
\end{qxbench}

\end{document}